\documentclass[10pt,letterpaper]{article}
\usepackage{arxiv}
\renewcommand{\shorttitle}{CoNav-UAV: Cooperative Dual-Altitude Aerial Navigation}
\usepackage[T1]{fontenc}
\usepackage[dvipsnames]{xcolor}
\usepackage{graphicx}
\usepackage[numbers,sort&compress]{natbib}
\usepackage[hyphens]{url}
\usepackage[font=small,labelfont=bf]{caption}
\usepackage{subcaption}
\usepackage{microtype}
\usepackage[section]{placeins}
\usepackage{adjustbox}
\usepackage{etoolbox}
\BeforeBeginEnvironment{tabular}{\begin{adjustbox}{max width=\linewidth}}
\AfterEndEnvironment{tabular}{\end{adjustbox}}
\date{}
\usepackage{algorithm}
\usepackage{algorithmic}
\usepackage{booktabs}
\usepackage{amsfonts}
\usepackage{amsmath}
\usepackage{amssymb}
\usepackage{mathtools}
\usepackage{amsthm}
\usepackage{multirow}
\usepackage{colortbl}
\usepackage{tabularx}
\usepackage{float}
\usepackage{enumitem}
\usepackage{pifont}

\usepackage{pgfplots}
\pgfplotsset{compat=1.18}
\usepgfplotslibrary{groupplots}
\usetikzlibrary{positioning,backgrounds,calc}
\definecolor{conavbg}{RGB}{232,242,252}
\definecolor{scenebg}{gray}{0.92}
\definecolor{ftblue}{HTML}{0173B2}
\definecolor{ftorange}{HTML}{DE8F05}
\definecolor{cvA}{HTML}{2A78D6}
\definecolor{cvB}{HTML}{EB6834}
\definecolor{cvC}{HTML}{1BAF7A}
\definecolor{cvD}{HTML}{EDA100}
\definecolor{cvMuted}{HTML}{8A8985}
\definecolor{cvGrid}{HTML}{E4E4E1}
\pgfplotsset{
  conav/.style={
    axis x line*=bottom, axis y line*=left,
    axis line style={black!70, line width=0.6pt},
    tick style={black!70, line width=0.6pt},
    tick align=outside, major tick length=2.5pt,
    xlabel near ticks, ylabel near ticks,
    tick label style={font=\sffamily\fontsize{8}{9.5}\selectfont, text=black!85},
    label style={font=\sffamily\fontsize{8}{9.5}\selectfont, text=black},
    title style={font=\sffamily\bfseries\fontsize{9}{11}\selectfont, text=black, yshift=-2pt},
    /pgf/number format/assume math mode=true,
    ymajorgrids, xmajorgrids=false,
    grid style={black!13, line width=0.5pt},
    scaled y ticks=false,
    legend style={draw=none, fill=none, font=\sffamily\fontsize{8}{9.5}\selectfont},
    every axis plot/.append style={line width=1.6pt, mark size=2.4pt,
      line join=round, line cap=round,
      mark options={solid, draw=white, line width=0.5pt}},
  },
}

\definecolor{cvprblue}{rgb}{0.21,0.49,0.74}
\usepackage[breaklinks,colorlinks,allcolors=cvprblue]{hyperref}

\DeclareMathOperator*{\argmax}{arg\,max}

\theoremstyle{plain}
\newtheorem{theorem}{Theorem}[section]
\newtheorem{proposition}[theorem]{Proposition}
\newtheorem{lemma}[theorem]{Lemma}
\newtheorem{corollary}[theorem]{Corollary}
\theoremstyle{definition}
\newtheorem{definition}[theorem]{Definition}

\theoremstyle{remark}
\newtheorem{remark}[theorem]{Remark}

\newcommand{\rev}[1]{#1}
\newcommand{\sd}[1]{\hspace{0.6pt}{\tiny\color{gray}(#1)}}
\newcommand{\dgood}[1]{{\scriptsize\color{teal}#1}}
\newcommand{\dbad}[1]{{\scriptsize\color{red}#1}}
\newcolumntype{Q}{>{\hsize=1.4\hsize\centering\arraybackslash}X<{\kern1.6em}}
\newcolumntype{Z}{>{\hsize=.8\hsize\centering\arraybackslash}X}

\newcommand{\revc}{}
\newcommand{\revb}[1]{#1}

\newcommand{\Tq}{\mathcal{T}}

\title{CoNav-UAV: Cooperative Dual-Altitude \\ Aerial Navigation via Stackelberg Learning}

\author{%
\begin{minipage}{0.96\textwidth}
\centering\normalfont\normalsize
Junru Song\textsuperscript{1,*}, Wenhao Zhang\textsuperscript{1,*},
Yang Yang\textsuperscript{2,*}, Xuekai Qiu\textsuperscript{2},\\[3pt]
Feifei Wang\textsuperscript{3}, Weien Zhou\textsuperscript{2},
Tingsong Jiang\textsuperscript{2}, Ying Wen\textsuperscript{1,4},\\[3pt]
Yang Li\textsuperscript{1,\dag}, Wen Yao\textsuperscript{2,\dag}\\[9pt]
\small
\textsuperscript{1}Shanghai Jiao Tong University\\
\textsuperscript{2}Intelligent Game and Decision Laboratory\\
\textsuperscript{3}Renmin University of China\quad
\textsuperscript{4}Shanghai Innovation Institute\\[5pt]
\textsuperscript{*}Equal contribution\quad
\textsuperscript{\dag}Corresponding authors
\end{minipage}}

\begin{document}
\maketitle

\begin{abstract}
{\revc
Aerial search missions often require an unmanned aerial vehicle (UAV) to locate a target from only a brief description of its appearance and surroundings, a task known as target-oriented aerial vision-and-language navigation (VLN).
The task demands city-scale exploration and grounding as well as collision-free close-range approach, two interleaved processes that are difficult to reconcile within a single agent.
Most existing methods adapt the ground VLN paradigm to a single UAV and rely on external assistance to compensate for its inefficiency.
A recent attempt deploys two UAVs at complementary altitudes but still depends on privileged sensing and trains the two agents independently, precluding any mutual adaptation.
We propose CoNav-UAV, which models the task as a Stackelberg game between a high-altitude leader and a low-altitude follower to capture their sequential decisions and asymmetric observations and utilities, with both agents using only onboard sensing.
We further introduce Iterative Stackelberg Learning, which tailors optimization to each agent's role: the leader's high-level vision-language reasoning builds on pretrained capabilities and is refined through memory-based in-context learning, while the follower's precise motion control is fine-tuned through DAgger-style expert distillation.
Alternating the two updates lets each agent continually adapt to the other's evolving behavior.
CoNav-UAV outperforms single- and dual-agent baselines across three high-fidelity urban scenes from the AerialVLN benchmark.
It improves success rate by up to 30.8 points on the learning scene and by up to 9.0 points under cross-scene transfer, where the leader directly reuses its learned memory and the follower is adapted with 2.6--3.7$\times$ less target-scene data.
Extensive ablations confirm the contributions of key components and the benefit of mutual adaptation, with consistent gains across VLM backbones.}
\end{abstract}

\section{Introduction}
\label{sec:intro}

\rev{Vision-and-language navigation (VLN)~\citep{anderson2018vln} requires an agent to reach a language-specified goal by coupling language understanding, visual grounding, and sequential decision making~\citep{chen2021hamt,an2024etpnav,zhou2024navgpt}.
Extending VLN to unmanned aerial vehicles (UAVs) has attracted growing interest~\citep{liu2023aerialvln,fan2023avdn,lee2024citynav,openfly2025,airnav2025}, yet aerial VLN remains considerably more challenging and less explored.
Navigation unfolds over city-scale scenes with trajectories spanning hundreds of meters~\citep{liu2023aerialvln,openfly2025}, which places far greater demands on exploration, target grounding, and long-horizon planning.
Practical missions such as disaster rescue, infrastructure inspection, and security patrol also rarely admit step-by-step route instructions, and operators can typically provide only a concise \emph{target-oriented} description of the target's appearance and surroundings~\citep{lee2024citynav,xu2025geonav,xiao2025uavon,zeng2026airgroundvln}.}

\rev{Most existing methods transfer the single-agent paradigm of ground VLN to a low-altitude UAV, which is confined to local observations and therefore explores inefficiently.
To compensate, they resort to external assistance such as human dialogue~\citep{fan2023avdn}, step-by-step route descriptions~\citep{liu2023aerialvln,openfly2025,liu2026openflyv2}, privileged directional hints~\citep{xu2026aerialvla}, real-time assistant guidance~\citep{wang2025traveluav}, and external landmark priors~\citep{xu2025geonav}.
Such reliance on privileged or human-provided information limits their practical applicability.}

\rev{A natural alternative is to pair a high-altitude UAV for wide-area search with a low-altitude UAV for close-range approach.
To our knowledge, AeroDuo~\citep{wu2025aeroduo} is the only prior attempt in this direction.
Despite substantial gains over single-UAV baselines, it has two critical limitations.
First, it still relies on privileged information, as its high-altitude UAV is steered by ground-truth target coordinates instead of exploring autonomously, and its low-altitude UAV navigates from point clouds and depth maps.
Second, its two agents are trained independently and composed only at inference, which leaves their interaction unmodeled and precludes mutual adaptation.
To date, no principled framework formalizes dual-altitude cooperation between agents with distinct utilities and jointly optimizes them from onboard information alone.}


\rev{To this end, we propose \textbf{CoNav-UAV}, a dual-altitude cooperative framework for target-oriented aerial VLN.
CoNav-UAV decomposes the task along the complementary observations of the two altitudes.
The high-altitude \emph{leader} is a frozen vision-language model (VLM) that exploits its wide field of view to explore the scene and ground candidate targets in bird's-eye-view (BEV) imagery.
The low-altitude \emph{follower} is a vision-language-action (VLA) model that exploits its close-range view to reach each dispatched target from first-person-view (FPV) images without collision.
The two agents communicate only through a lightweight coordinate queue, without shared perception or privileged sensing.
We formulate their cooperation as a general-sum Stackelberg game~\citep{von2010market}, which captures their sequential decisions and their asymmetric observations and utilities.
This formulation makes the mutual dependence between the two agents explicit, which motivates our solution method and offers a formal view of its learning dynamics.}

\rev{To solve this game, we develop \emph{Iterative Stackelberg Learning} (ISL), which tailors optimization to each agent's role.
The leader's vision-language reasoning builds on the pretrained capabilities of the VLM and is refined without gradients through episodic memory with value-aware retrieval~\citep{zhang2026memrl}.
The follower's motion control is fine-tuned through DAgger-style distillation~\citep{ross2011reduction} from a compact PPO expert~\citep{schulman2017proximal}.
The two updates alternate so that each agent optimizes its own utility while adapting to the evolving behavior of the other, which leads to their joint improvement.}

\rev{Experiments on three high-fidelity urban scenes from the AerialVLN benchmark~\citep{liu2023aerialvln} show that CoNav-UAV outperforms single- and dual-agent baselines on all three scenes and attains the highest success rate in eight of the nine scene--difficulty settings.
It improves the success rate by up to 30.8 points on the learning scene and by up to 9.0 points in cross-scene transfer, where the leader directly reuses its learned memory and the follower is adapted with 2.6--3.7$\times$ less target-scene data.
Ablation studies further confirm the contributions of the key components and the benefit of mutual adaptation, with consistent gains across VLM backbones.
\revb{To support future research, we will release our code, the data generation pipeline, and the high- and low-altitude data used for learning and evaluation.}}

\rev{In summary, our contributions are threefold.
(i)~We present, to our knowledge, the first game-theoretic formulation of dual-altitude UAV cooperation, cast as a general-sum Stackelberg game.
(ii)~We propose Iterative Stackelberg Learning, which tailors optimization to each agent's role and achieves mutual adaptation through alternating updates.
(iii)~Experiments in photorealistic simulation demonstrate clear gains over strong baselines, efficient cross-scene transfer, and robustness across VLM backbones.}

\begin{figure}[H]
\centering
\includegraphics[width=0.99\textwidth]{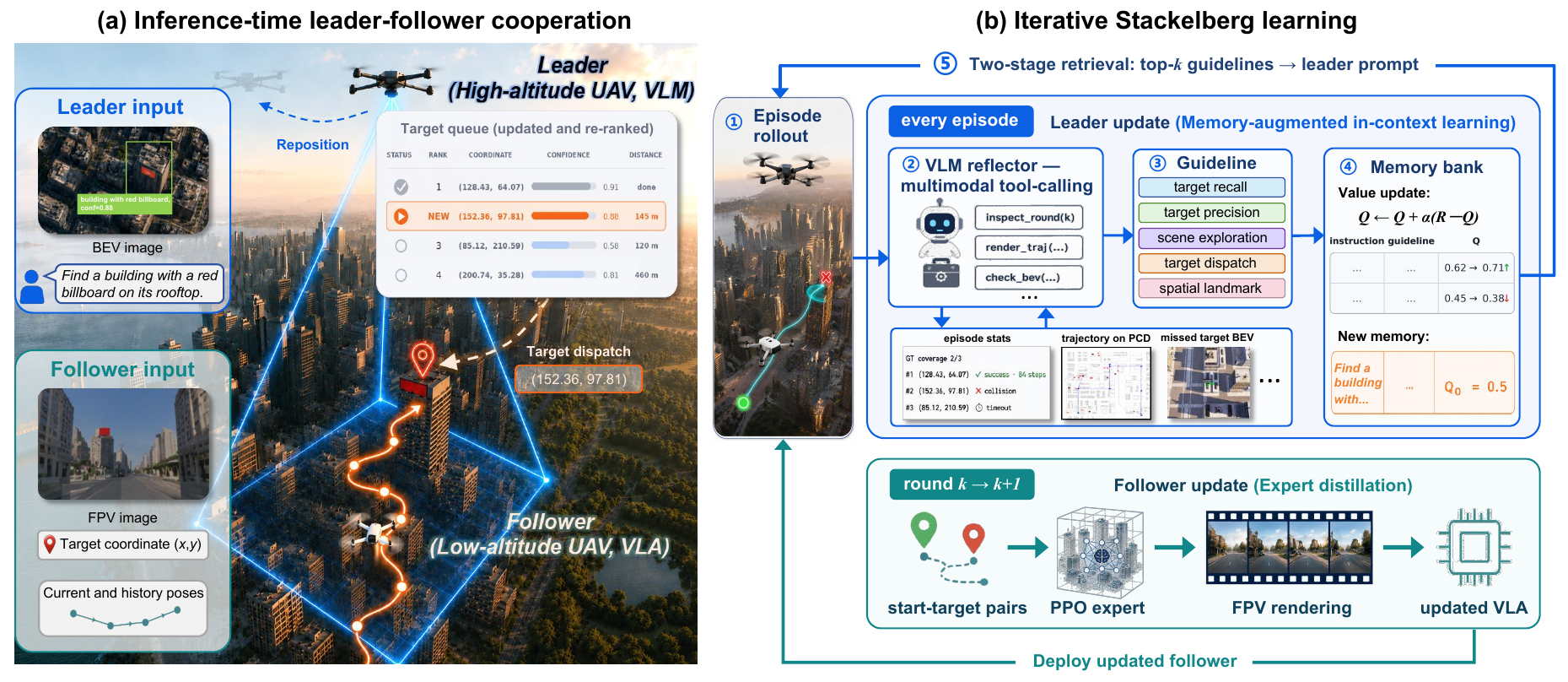}
\caption{%
  Overview of CoNav-UAV.
  \textbf{(a)}~Inference-time cooperation: the leader performs BEV
  grounding and maintains a target queue; the follower navigates to
  dispatched targets \rev{from FPV images}.
  \textbf{(b)}~Iterative Stackelberg Learning: the leader update
  evolves the memory bank via multimodal reflection and
  Q-value learning per episode; the follower update fine-tunes the
  VLA via DAgger distillation from a PPO expert per round.%
}
\label{fig:overview}
\end{figure}

\section{Related Work}
\label{sec:related}

\subsection{Aerial Vision-and-Language Navigation}
\label{sec:related_vln}

\rev{Aerial VLN has advanced rapidly with the introduction of dedicated benchmarks~\citep{fan2023avdn,liu2023aerialvln,lee2024citynav,openfly2025,airnav2025}.
Most methods follow step-by-step route instructions under the single-agent paradigm of ground VLN~\citep{anderson2018vln,krantz2020vlnce}, and have advanced from sequence-to-sequence and cross-modal attention models~\citep{liu2023aerialvln,fan2023avdn,lee2024citynav} to LLM- and VLM-based agents~\citep{zhang2025citynavagent,gao2024stmrvln,hu2025spf} and end-to-end VLA models~\citep{xu2026aerialvla,openfly2025,liu2026openflyv2}.
The more practical target-oriented setting provides only a concise description of the target~\citep{lee2024citynav,wang2025traveluav,xiao2025uavon,pardyl2025flysearch,zeng2026airgroundvln}, under which a single UAV confined to egocentric observations searches inefficiently and typically depends on the external assistance listed in Sec.~\ref{sec:intro}.
APEX~\citep{zhang2026apex} avoids such assistance by maintaining a VLM-built spatio-semantic memory of the explored area, but it remains a single-UAV system that must reconcile wide-area search and close-range approach within one agent.}

\rev{Only a few recent systems explore cooperation between heterogeneous agents for this task.
AGC-VLN~\citep{zhang2026agcvln} connects a UAV and a ground vehicle through a shared bird's-eye map on which the target is marked at its known position, and AeroDuo~\citep{wu2025aeroduo} steers its high-altitude UAV by the target position while its low-altitude UAV executes waypoints from privileged geometric inputs.
Both assume access to the target position and join separately built agents through a fixed interface, so the agents cannot adapt to each other.
AG-CoNAV~\citep{zeng2026airgroundvln} instead coordinates an aerial and a ground platform within a single model, which requires reference trajectories for both platforms, depth sensing, and continuous sharing of their observations.
In contrast, CoNav-UAV keeps two separate agents that exchange only target coordinates, and its leader discovers and grounds targets autonomously by adapting a pretrained VLM in context.
More fundamentally, CoNav-UAV treats the cooperation itself as the object of learning, modeling the leader--follower interaction as a Stackelberg game and optimizing each agent against the other.}

\subsection{Learning in Stackelberg Games}
\label{sec:related_stackelberg}

\rev{Stackelberg learning seeks a Stackelberg equilibrium (SE), in which the leader optimizes in anticipation of the follower's best response, and has been extended to Markov games along several lines.
Existing methods require the follower's objective to be differentiable in the leader's parameters~\citep{fiez2020implicit,goktas2023convex,zhong2025bilevel}, rely on linear function approximation~\citep{chen2025exploration}, or assume known follower objectives and dynamics~\citep{li2025neural} or query access to the follower's best response~\citep{gerstgrasser2023oracles}.
None of these assumptions holds in our setting, where the leader is a frozen VLM adapted through natural-language guidelines and the follower is a large VLA whose best response can be neither differentiated nor queried.
The formulation closest to ours casts model-based RL as a Stackelberg game between a policy and a learned dynamics model~\citep{rajeswaran2020game}.
Since the model is accurate only on the state distribution induced by the current policy, the two players are updated alternately.
Our follower is in an analogous position, as its limited capability requires fitting it to the dispatches of the current leader, and we adopt the same alternating scheme.}

\section{\rev{Dual-Altitude Cooperation as a Stackelberg Game}}
\label{sec:formulation}

\subsection{Task and Agent Roles}
\label{sec:roles}

\rev{The task is specified by a target-oriented natural-language instruction~$\ell$, which describes the target's visual attributes, surrounding landmarks, and approximate location but provides no step-by-step route.
Both UAVs rely solely on onboard RGB cameras and their own poses, which are standard on commodity UAVs.
The VLM-based leader grounds candidates matching~$\ell$ in BEV imagery into a target queue~$\Tq$ prioritized by detection confidence and path efficiency, repositions itself to explore unsurveyed regions or to confirm candidates at closer range, and decides when to terminate the mission.
The VLA-based follower flies at a \rev{lower} altitude and navigates to the head of~$\Tq$ from FPV images while avoiding collisions, so the leader's dispatches define its task.}

\rev{This role division follows the complementary observations at the two altitudes and also aligns with the cost of supervision.
Instruction understanding, target recognition, and exploration require costly annotations and are therefore delegated to a frozen VLM that improves through in-context learning (Sec.~\ref{sec:leader}), whereas close-range navigation is learned from instruction-free point-goal trajectories that are far cheaper to obtain (Sec.~\ref{sec:follower}).
Supp.~\ref{app:scope} discusses \rev{how this design extends to other follower embodiments}.}

\subsection{\rev{Game Formulation}}
\label{sec:game}
\label{sec:prelim}

\rev{In a Stackelberg game~\citep{von2010market}, a \emph{leader} first commits to an action, and a \emph{follower} observes it and best-responds, while the leader optimizes in anticipation of this response.
Dual-altitude cooperation shares this structure, since the leader's dispatches precede and define the follower's task (\emph{sequential commitment}), are chosen in anticipation of the targets that the follower can reach (\emph{anticipatory optimization}), and serve a different objective (\emph{distinct utilities}).}

\begin{definition}[Dual-Altitude Stackelberg Navigation Game]
\label{def:game}
\rev{The game is a tuple
$\mathcal{G}=\langle \mathcal{S},\,\ell,\,
\mathcal{O}_L,\mathcal{O}_F,\,
\mathcal{A}_L,\mathcal{A}_F,\,
P,\, r^L,\, r^F,\, \gamma,\, m\rangle$
with state space $\mathcal{S}$, target-oriented instruction $\ell$, observation and action spaces $\mathcal{O}_i$ and $\mathcal{A}_i$ of agent $i\!\in\!\{L,F\}$, transition kernel $P$, reward functions $r^L$ and $r^F$, discount factor $\gamma$, and leader decision interval $m$.
Each instruction induces one game instance and determines the ground-truth target set.
At each decision time $\tau_k\!:=\!km$, $k=0,\ldots,H{-}1$, the leader observes $o^L_k\!\in\!\mathcal{O}_L$ and commits to an action $a^L_k\!\in\!\mathcal{A}_L$ that dispatches candidate targets and relocates its viewpoint.
Over the following $m$ steps, the follower receives the dispatched target through $o^F_t\!\in\!\mathcal{O}_F$ and responds with motion actions $a^F_t\!\in\!\mathcal{A}_F$.}
\end{definition}

\paragraph{\rev{State, observations, and actions.}}
\rev{The state
$s_t=(\mathcal{E},\,\ell,\,p^L_t,\,p^F_t,\,\Tq_t,\,\mathcal{H}_t)$
comprises the 3-D scene $\mathcal{E}$, the instruction, the UAV poses $p^L_t$ and $p^F_t$, the target queue $\Tq_t$, and a record $\mathcal{H}_t$ of past detections and dispatch outcomes.
Each agent receives a partial observation}
\begin{equation}
  o^L_k = \bigl(I^{\mathrm{bev}}_k,\ell,
    p^L_{\tau_k},p^F_{\tau_k},\Tq_{\tau_k},
    \mathcal{H}_{\tau_k}\bigr),
  \;\;
  o^F_t = \bigl(I^{\mathrm{fpv}}_t,
    c^{\mathrm{tgt}}_t,p^F_t,h^F_t\bigr),
  \label{eq:obs}
\end{equation}
\rev{where $I^{\mathrm{bev}}_k$ and $I^{\mathrm{fpv}}_t$ are the BEV and FPV images, $c^{\mathrm{tgt}}_t\!\in\!\mathbb{R}^2$ is the active target coordinate, and $h^F_t\!=\!(p^F_{t-1},\ldots,p^F_{t-w})$ is the history of the follower's $w$ previous poses.
At each decision step, the leader updates $\Tq$ and either repositions its viewpoint by a fixed-step horizontal translation or a discrete altitude change, or terminates the mission.
The follower outputs a planar displacement and a yaw adjustment $a^F_t\!=\!(\Delta x,\Delta y,\Delta\psi)\!\in\!\mathbb{R}^3$ at each step.}

\paragraph{Rewards \rev{and utilities}.}
\rev{Let $\mathcal{C}^\star\!=\!\{c_1^\star,\ldots,c_{N^\star}^\star\}$ denote the ground-truth (GT) targets satisfying~$\ell$.
Since the leader's decisions take effect only after the follower navigates, the leader receives a single \emph{sparse} reward at termination ($t\!=\!T$)}
\begin{equation}
  \label{eq:rL}
  r^L_T = \frac{N_{\mathrm{hit}}}{N^\star}
        - \sum_{q\in\mathcal{P}}
          \alpha_q\,n_q,
  \qquad
  r^L_t = 0 \;\;\text{for}\;\; t<T,
\end{equation}
\rev{where $N_{\mathrm{hit}}/N^\star$ is the GT recall and $\mathcal{P}\!=\!\{\mathrm{fp},\mathrm{coll},\mathrm{waste},\mathrm{osc}\}$ indexes four penalized events, namely false-positive dispatches, collision endings, wasted grounding rounds, and oscillations, with counts $n_q$ and weights $\alpha_q\!>\!0$.
Collisions are charged to the leader, since dispatching an unreachable or hazardous target is a scheduling failure.
The leader's utility is its expected discounted return $J^L(\pi_L,\pi_F)=\mathbb{E}_{s_0,\pi_L,\pi_F}\bigl[\sum_{t=0}^{T}\gamma^t r^L_t\bigr]$.
The follower receives a dense, step-level reward that combines target approach and collision avoidance}
\begin{equation}
  \label{eq:rF}
  r^F_t = \beta_p\,\Delta d_t
        - \beta_c\,\mathbf{1}[\mathrm{coll}_t],
\end{equation}
\rev{where $\Delta d_t\!:=\!d_{t-1}\!-\!d_t$ is the reduction in distance to the active target and $\mathrm{coll}_t$ indicates obstacle proximity below a threshold~$\delta$.
Unlike $r^L_T$, this reward depends only on the active target, so the follower's utility is defined per dispatch from follower state $x$ to target $c$ as}
{\revc
\begin{equation}
  \label{eq:jf}
  J^F(\pi_F;x,c)=\mathbb{E}\Bigl[\textstyle\sum_{t<T_c}\gamma^t r^F_t\Bigr],
\end{equation}}%
\rev{where $T_c$ is the step of arrival or collision.}

\begin{definition}[Stackelberg Equilibrium, SE]
\label{def:se}
\rev{Let $\Pi_L$ and $\Pi_F$ denote the sets of policies attainable by the leader and the follower, and let $d_{\pi_L}$ denote the distribution of dispatches $(x,c)$ induced by $\pi_L$.
The follower's best response to a leader policy $\pi_L$ is}
\begin{equation}
  \label{eq:br}
  \mathrm{BR}(\pi_L)
  :=\argmax_{\pi_F\in\Pi_F} {\revc\mathbb{E}_{(x,c)\sim d_{\pi_L}} J^F(\pi_F;x,c)}.
\end{equation}
\rev{A policy pair $(\pi_L^\star,\pi_F^\star)$ is a Stackelberg equilibrium of $\mathcal{G}$ if}
\begin{equation}
  \label{eq:se}
  \pi_L^\star
  \in\argmax_{\pi_L\in\Pi_L}
    J^L\bigl(\pi_L,\mathrm{BR}(\pi_L)\bigr),
  \quad
  \pi_F^\star=\mathrm{BR}(\pi_L^\star).
\end{equation}
\end{definition}

\paragraph{\rev{Solving the game with a capacity-limited follower.}}
\rev{We adopt the SE as the solution concept of $\mathcal{G}$.
Since $J^F$ depends only on the dispatch, an ideal follower would admit a single reaction function that is optimal for every dispatch $(x,c)$ and independent of the leader, as in the classical model~\citep{von2010market}.
The deployed VLA, however, navigates from partial onboard observations with limited capability, so it cannot realize this reaction function and must instead be fitted to the dispatches of a specific leader.
Its best response thus depends on the leader's dispatch distribution $d_{\pi_L}$, while the leader's optimal dispatches depend in turn on the targets that this follower can reach (Propositions~\ref{prop:reaction} and~\ref{prop:mutual}, Supp.~\ref{app:theory}).
Since recomputing the follower's best response for every candidate leader is intractable, we solve the game iteratively, updating each agent against the latest policy of the other and using the current follower as a surrogate for its best response to the updated leader~\citep{rajeswaran2020game}.}

\section{Iterative Stackelberg Learning}
\label{sec:method}

\subsection{Overview}
\label{sec:overview}

\rev{\emph{Iterative Stackelberg Learning} (ISL) approximates the bilevel structure of the SE (Definition~\ref{def:se}) by alternating two updates, each tailored to the role of its agent (Fig.~\ref{fig:overview}(b)).
The leader update approximates the leader's optimum against the current follower.
Since the leader is a frozen VLM whose parameters and gradients are inaccessible, this update is performed in context by conditioning the leader policy on natural-language \emph{guidelines} retrieved from an episodic memory~\citep{zhang2026memrl} (Sec.~\ref{sec:leader}).
The follower update in turn approximates the follower's best response by expert distillation on the leader's dispatches~\citep{ross2011reduction} (Sec.~\ref{sec:follower}).}

\rev{We define a \emph{round} as one full pass over the learning episodes.
The leader update is gradient-free and revises the memory after every episode, whereas the follower update requires VLA fine-tuning and thus runs once per round.
Within a round, the leader adapts to the current capability of the fixed follower.
The follower update then re-aligns the follower with the leader's evolved dispatch behavior, which keeps its training distribution matched to deployment.
Algorithm~\ref{alg:main} in Supp.~\ref{app:algorithm} summarizes the full procedure, which starts from an empty memory bank and a bootstrap follower $\pi_F^{(0)}$.}

\subsection{Leader Update: Memory-Augmented In-Context Learning}
\label{sec:leader}

\rev{The VLM weights remain frozen, and the leader improves entirely through its prompt via in-context learning (ICL) over an episodic memory~\citep{zhang2026memrl}.
Each entry of the memory bank
$\mathcal{M}=\{(\ell_i,e_i,Q_i)\}_{i=1}^{|\mathcal{M}|}$
comprises an instruction key~$\ell_i$, an experience guideline~$e_i$, and a learned utility estimate~$Q_i$.}

\paragraph{Rollout, reflection, and guideline generation.}
\rev{Upon completion of an episode, the outcomes of the dispatched targets and notable events such as collisions are annotated automatically.
A VLM reflector then analyzes the episode and distills it into a structured guideline}
\begin{equation}
  \label{eq:reflect}
  e_j = \mathrm{Reflect}(\mathrm{ep}_j).
\end{equation}
\rev{Instead of ingesting the full episode trace, the reflector iteratively invokes multimodal analysis tools and selects which grounding rounds, navigation segments, BEV images, or other artifacts to inspect at each step.
Such on-demand inspection avoids diluting critical information in a lengthy context and focuses the reasoning on the most informative evidence (example traces in Supp.~\ref{app:qual_reflection}).
Both successful and failed episodes yield guidelines in a common format, the former contributing reusable strategies and the latter diagnostic recommendations.
Each guideline covers five dimensions, namely target recall, precision, scene exploration, target dispatch, and spatial landmarks, which are aligned with the components of $r^L_T$ (Eq.~\ref{eq:rL}) so that in-context adaptation addresses the capabilities that the reward evaluates.}

\paragraph{Memory writing and utility learning.}
\rev{For a first-seen instruction, $(\ell_j,e_j)$ is stored as a new entry, and otherwise the existing guideline is revised.
A new entry is initialized with $Q_0^{+}$ if the episode attains at least 50\% GT coverage and with a lower value $Q_0^{-}$ otherwise (values in Supp.~\ref{app:leader}), which biases early retrieval toward validated strategies.
After each episode with reward $R\!=\!r^L_T$, the Q-values of all retrieved entries are updated by an exponential moving average}
\begin{equation}
  \label{eq:q_update}
  Q_i \leftarrow Q_i + \eta\,(R - Q_i),
  \quad \forall\; i\in\mathcal{M}_{\mathrm{res}},
\end{equation}
\rev{with learning rate $\eta\!\in\!(0,1)$.
Under stationary conditions, $Q_i$ converges to the expected leader reward obtained when entry~$i$ is retrieved~\citep{zhang2026memrl}.}

\paragraph{Value-aware retrieval.}
\rev{Given the next instruction~$\ell$, entries are selected in two phases.
First, embedding similarity between~$\ell$ and each stored key~$\ell_i$ yields a pool $\mathcal{C}(\ell)$ of the top-$k_1$ candidates.
Second, a composite score balances semantic similarity against learned utility within this pool}
\begin{equation}
  \label{eq:retrieve}
  \mathrm{score}_i
  =(1{-}\lambda)\,\hat{s}_i
  +\lambda\,\hat{Q}_i,
\end{equation}
\rev{where $\hat{s}_i$ and $\hat{Q}_i$ are the z-score-normalized similarity and Q-value, and $\lambda\!\in\![0,1]$ controls their balance.
The top-$k_2$ entries form $\mathcal{M}_{\mathrm{res}}(\ell)$, with $\varepsilon$-greedy exploration to prevent premature convergence.
These guidelines are then injected into the VLM prompt, which conditions the leader policy $\pi_L(\cdot\mid o^L;\mathcal{M}_{\mathrm{res}}(\ell))$ and closes the per-episode loop.}

\subsection{Follower Update: Expert Distillation}
\label{sec:follower}

\rev{Coordinate-conditioned collision-free navigation demands fine-grained control, yet end-to-end RL on a large VLA is prohibitively sample-intensive.
We therefore train a lightweight RL expert in a geometric proxy environment and let the VLA imitate its actions from onboard observations through supervised fine-tuning (SFT) (Fig.~\ref{fig:overview}(b)).}

\paragraph{Start--target pairs.}
\rev{The initial follower $\pi_F^{(0)}$ is bootstrapped from randomly sampled start--target pairs, which yields a leader-agnostic policy with broad coverage.
In subsequent rounds, the pairs are instead drawn from the navigation segments of cooperative rollouts.
They thus track the leader's evolving dispatch distribution and concentrate the limited capacity of the VLA on the dispatches that the current leader actually issues.}

\paragraph{\rev{RL expert as action annotator.}}
\rev{The 3-D scene is represented as a point-cloud occupancy grid, which preserves collision geometry without rendering.
A compact CNN policy $\pi_E$ is trained with PPO~\citep{schulman2017proximal} on $r^F_t$ (Eq.~\ref{eq:rF}) to map a local occupancy patch and the target coordinate to a displacement command.
The expert observes the exact geometry around the UAV, which the VLA must infer from FPV images, and with this privileged observation it closely approximates the ideal reaction function of Sec.~\ref{sec:game}.
Since $r^F_t$ does not involve the leader, the same expert serves as the teacher in the bootstrap stage and in every DAgger round, where it annotates start--target pairs with actions.
We favor RL over classical planners for this role because the expert is trained end to end on the reward without hand-crafted heuristics, and labeling a new pair requires only an inexpensive rollout of the compact policy (Supp.~\ref{app:pipeline_low}).
Such automatic generation from scene point clouds follows established practice in aerial VLN~\citep{openfly2025}, and our role division simplifies it further because the follower's purely geometric task needs no subsequent instruction synthesis.
The same decoupling helps where no point cloud is available and demonstrations must be flown by human pilots, because a point-goal flight requires no language annotation and is therefore easier to collect than a conventional VLN trajectory.}

\paragraph{FPV rendering and distillation.}
\rev{Expert rollouts are replayed in the physics simulator to render synchronized FPV images.
Each image, together with the other components of $o^F_t$ (Eq.~\eqref{eq:obs}), forms the input of a training sample, and the expert action $a_t^\star$ serves as its label (prompt template in Supp.~\ref{app:vla}).
The VLA is initialized from Qwen2.5-VL-7B-Instruct~\citep{bai2025qwen2} and fine-tuned with LoRA~\citep{kim24openvla}, with each action dimension discretized into 256 autoregressively decoded bins and trained with token-level cross-entropy.
In round~$k$, the DAgger data $\mathcal{D}^{(k)}_{\mathrm{dag}}$ are added to the training set and the follower is retrained as $\pi_F^{(k+1)} \leftarrow \mathrm{SFT}(\pi_F^{(k)},\,\mathcal{D}^{(0)}_{\mathrm{init}}\cup\bigcup_{j\le k}\mathcal{D}^{(j)}_{\mathrm{dag}})$, where the bootstrap dataset $\mathcal{D}^{(0)}_{\mathrm{init}}$ is retained to prevent forgetting.
\rev{For onboard deployment, we also study smaller followers (Supp.~\ref{app:deployment}).
A compact network trained from scratch falls short at long range, which shows that the follower depends on the spatial priors of the pretrained VLM.
Distillation retains them at about one ninth of the parameters, with a comparable success rate.}}

\subsection{\rev{Relation to the Stackelberg Equilibrium}}
\label{sec:isl_se}

\rev{The Stackelberg formulation also offers a formal view of the learning dynamics of ISL.
Supp.~\ref{app:theory} decomposes the gap to the SE value into the leader's suboptimality against the current follower and the follower's deviation from its best response on the leader's dispatches (Corollary~\ref{cor:gap}), and the leader and follower updates reduce these two terms respectively.
The same decomposition explains the non-monotone iteration curve in Sec.~\ref{sec:iteration}.
Both agents are large models updated through in-context memory and supervised fine-tuning, and we leave a convergence analysis of the alternation to future work, together with a quantitative bound on the distance between its fixed points and the SE.}

\section{Experiments}
\label{sec:experiments}

\rev{We compare CoNav-UAV with baselines (Sec.~\ref{sec:main_results}), trace its improvement across Stackelberg rounds (Sec.~\ref{sec:iteration}), and ablate its design (Sec.~\ref{sec:ablation_studies}).
Supp.~\ref{app:leader_quality} further analyzes the leader's grounding quality \rev{across six VLM backbones}.}

\subsection{Experimental Setup}
\label{sec:setup}

\paragraph{Simulation environment.}
All experiments are conducted in AirSim-based photorealistic
environments rendered with Unreal Engine on the OpenFly
platform~\citep{openfly2025}, using three scenes from the AerialVLN
dataset~\citep{liu2023aerialvln}, namely \texttt{airsim16}
(3.7\,km$^2$), \texttt{airsim26} (9.0\,km$^2$), and
\texttt{airsim23} (0.62\,km$^2$), spanning diverse architectural styles
and densities.
\texttt{airsim16} serves as the learning scene, and the other two
are held out for transfer.

\paragraph{Benchmark dataset.}
Ground-truth targets and instructions are annotated in-simulator
through a two-stage labeling-then-verification pipeline
(Supp.~\ref{app:pipeline_high}),
and randomly split into two thirds for learning and one third for
held-out testing.
Each task instance is a long-horizon search episode \rev{with one or more targets} that
spans hundreds of meters.
All results are averaged over 3 seeded runs\rev{, and all reported $p$-values come from one-sided Welch's $t$-tests}.
 Benchmark statistics are given in Supp.~\ref{app:data_release}.

\paragraph{Evaluation metrics.}
\rev{Target-oriented navigation is exploratory and has no predetermined endpoint, so we replace single-endpoint metrics~\citep{anderson2018vln} with the trajectory-level reachability of GT targets under a fixed budget.}
\textbf{NE$_{\min}$} (Minimum Navigation Error, $\downarrow$) is
the minimum distance between the trajectory and each GT target,
averaged over targets.
\textbf{OSR} (Oracle Success Rate, $\uparrow$) is the fraction of
GT targets with NE$_{\min}\!\le\!\varepsilon$
($\varepsilon{=}20$\,m), computed as recall over targets.
\textbf{SPL$_{\mathrm{OSR}}$} (Success weighted by Path Length,
$\uparrow$) is OSR weighted by $L^*/\max(L^*,P)$, where $L^*$ is
the greedy path through all GT targets and $P$ is the trajectory
length.
The metrics accommodate multi-target instructions and apply
symmetrically to single- and multi-agent methods.
A trajectory terminates upon collision, full target coverage, or
budget exhaustion ($B\!=\!3L^*$).
Test episodes are partitioned by start-to-target distance into
\textbf{Easy} (100\,m), \textbf{Medium} (200\,m), and
\textbf{Hard} (300\,m) tiers, where larger distances demand more exploration
and longer flights.

\paragraph{Implementation.}
The follower VLA is initialized from Qwen2.5-VL-7B-Instruct via
SFT on 6K expert trajectories from the PPO pipeline and updated by
one DAgger epoch per round.
When transferring to a new scene, it is further adapted with
$\sim$2K trajectories from that scene while the leader memory is
carried over unchanged; adaptation details and corpus statistics
are provided in Supp.~\ref{app:vla} and~\ref{app:data_release}.
As leader we evaluate four frozen VLMs, the proprietary
GPT-5.6-sol and GPT-5.4-mini and the open-source
Qwen3-VL-32B/8B-Instruct, assisted by GroundingDINO for candidate
detection on BEV imagery (details in Supp.~\ref{app:leader}).

\paragraph{Baselines.}
\rev{\textbf{CMA}~\citep{tan2019lxmert} and \textbf{Seq2Seq}~\citep{sutskever2014seq2seq} are lightweight single-agent models widely used in aerial VLN benchmarks, \textbf{OpenFly}~\citep{openfly2025} and \textbf{AerialVLA}~\citep{xu2026aerialvla} are VLA-based single-agent navigators, and \textbf{AeroDuo}~\citep{wu2025aeroduo} is the only prior aerial method that pairs UAVs at two altitudes.}
All baselines are evaluated under the same protocol, with
implementation details in Supp.~\ref{app:baseline_impl}.

\subsection{Main Results}
\label{sec:main_results}

\paragraph{Comparison with Baselines.}

\emph{(1)~CoNav-UAV is \rev{consistently} stronger in-distribution.}
In Table~\ref{tab:main}, the strongest configuration per tier reaches 40.6\% OSR on Easy
(GPT-5.4-mini), 35.7\% on Medium, and 37.2\% on Hard (both
GPT-5.6-sol), exceeding the strongest baseline in each tier by
+14.7, +26.4, and +30.8\,pp, respectively.

\emph{(2)~The advantage transfers across scenes at a much lower
adaptation cost.}
CoNav-UAV attains the highest OSR in five of the six cross-scene
tier combinations using leader memory transferred from
\texttt{airsim16}.
On \texttt{airsim26} Hard, the strongest CoNav-UAV result is only
0.7\,pp below AerialVLA (15.2\% vs.\ 15.9\%); on
\texttt{airsim23}, CoNav-UAV leads the strongest baseline by
6.3, 9.0, and 6.9\,pp from Easy to Hard.
The baselines reach these numbers only after fine-tuning on
target-scene data with new instruction annotations, 4.6K
trajectories on \texttt{airsim23} and 7.5K on \texttt{airsim26},
whereas CoNav-UAV adapts its follower with about 2K
\rev{instruction-free} trajectories, $2.6\times$ and $3.7\times$ fewer,
and leaves the leader untouched
(Supp.~\ref{app:baseline_impl}).

\emph{(3)~Long-range target-oriented navigation remains the main
source of difficulty.}
Baseline OSR generally falls with distance, while the dual-altitude system preserves a larger fraction of its Easy-tier coverage.
\rev{The remaining gap on \texttt{airsim26} Hard nevertheless shows that long-range grounding and control stay challenging after transfer.}

\rev{Against the strongest baseline in each scene, the average over the four leaders is significantly better on \texttt{airsim16} ($p<0.001$) and \texttt{airsim23} ($p=0.004$) and marginally better on \texttt{airsim26} ($p=0.09$).
The strongest leader, GPT-5.6-sol, is significantly better in all three scenes ($p<0.001$, $p=0.014$, and $p=0.025$, respectively).}

\rev{Supp.~\ref{app:fine_grained}, \ref{app:gen_full}, and~\ref{app:qual_rollout} provide per-VLM breakdowns, the remaining baseline variants, and qualitative rollouts.}

\begin{table}[H]
\caption{%
  Comparison with baselines on the learning scene \texttt{airsim16} and under cross-scene
  adaptation to \texttt{airsim26/23}, by difficulty tier.
  \rev{On \texttt{airsim16}, CoNav-UAV is evaluated after two rounds of ISL.}
  On \texttt{airsim26/23}, all baselines are fine-tuned on the target scene, while CoNav-UAV adapts its follower
  with about 2K trajectories per scene and directly reuses the leader
  memory learned on \texttt{airsim16}.
  Best and second-best are marked within each scene and tier.
  \rev{Gray parentheses give the standard deviation over runs.}%
}
\label{tab:main}
\centering\fontsize{8}{9.5}\selectfont
\setlength{\tabcolsep}{1.5pt}%
\renewcommand{\arraystretch}{0.90}
\begin{tabularx}{\textwidth}{@{}l *{9}{>{\centering\arraybackslash}X}@{}}
\toprule
& \multicolumn{3}{c}{Easy \rev{(100\,m)}}
& \multicolumn{3}{c}{Medium \rev{(200\,m)}}
& \multicolumn{3}{c}{Hard \rev{(300\,m)}} \\
\cmidrule(lr){2-4}\cmidrule(lr){5-7}\cmidrule(lr){8-10}
Method
  & OSR$\uparrow$ & NE$\downarrow$ & SPL$\uparrow$
  & OSR$\uparrow$ & NE$\downarrow$ & SPL$\uparrow$
  & OSR$\uparrow$ & NE$\downarrow$ & SPL$\uparrow$ \\
\midrule
\rowcolor{scenebg}
\multicolumn{10}{@{}l@{}}{%
  \textbf{\texttt{airsim16}} \textit{(learning scene)}} \\
CMA~\citep{tan2019lxmert}
  & 4.3\sd{3.2} & 64.6\sd{5.5} & .043\sd{.032}
  & 1.3\sd{2.2} & 152.8\sd{4.9} & .013\sd{.022}
  & 1.3\sd{2.2} & 234.7\sd{0.8} & .013\sd{.022} \\
Seq2Seq~\citep{sutskever2014seq2seq}
  & 2.6\sd{2.2} & 69.5\sd{3.9} & .019\sd{.019}
  & 0\sd{0.0} & 162.4\sd{4.2} & 0\sd{.000}
  & 0\sd{0.0} & 256.0\sd{3.6} & 0\sd{.000} \\
OpenFly~\citep{openfly2025}
  & 4.3\sd{0.7} & 71.8\sd{4.0} & .040\sd{.004}
  & 0\sd{0.0} & 167.1\sd{4.2} & 0\sd{.000}
  & 2.6\sd{2.2} & 253.6\sd{18.3} & .022\sd{.019} \\
AerialVLA~\citep{xu2026aerialvla}
  & 25.9\sd{5.8} & 63.0\sd{1.9} & .255\sd{.056}
  & 6.8\sd{4.1} & 148.2\sd{14.7} & .068\sd{.041}
  & 1.3\sd{2.2} & 231.2\sd{16.5} & .013\sd{.022} \\
AeroDuo~\citep{wu2025aeroduo}
  & 24.4\sd{8.5} & 74.8\sd{3.9} & .172\sd{.048}
  & 9.3\sd{3.9} & 142.9\sd{7.5} & .067\sd{.033}
  & 6.4\sd{3.1} & 280.6\sd{27.4} & .035\sd{.025} \\
\rowcolor{conavbg}
CoNav-UAV (GPT-5.6-sol)
  & \underline{36.4}\sd{8.5} & \textbf{43.9}\sd{4.1} & \underline{.308}\sd{.059}
  & \textbf{35.7}\sd{5.1} & \textbf{73.1}\sd{6.2} & \textbf{.284}\sd{.058}
  & \textbf{37.2}\sd{5.4} & \textbf{95.1}\sd{12.4} & \textbf{.309}\sd{.038} \\
\rowcolor{conavbg}
CoNav-UAV (GPT-5.4-mini)
  & \textbf{40.6}\sd{9.8} & 45.0\sd{6.0} & \textbf{.334}\sd{.085}
  & \underline{34.6}\sd{3.9} & \underline{80.0}\sd{10.3} & \underline{.253}\sd{.054}
  & \underline{22.6}\sd{4.5} & 128.4\sd{21.6} & .146\sd{.021} \\
\rowcolor{conavbg}
CoNav-UAV (Qwen3-VL-32B)
  & 32.6\sd{8.9} & 47.5\sd{4.6} & .280\sd{.079}
  & 25.1\sd{7.0} & 91.0\sd{6.9} & .208\sd{.055}
  & 21.3\sd{9.7} & 124.5\sd{15.7} & \underline{.168}\sd{.075} \\
\rowcolor{conavbg}
CoNav-UAV (Qwen3-VL-8B)
  & 26.5\sd{3.0} & \underline{44.8}\sd{1.4} & .225\sd{.015}
  & 18.0\sd{4.4} & 80.1\sd{6.5} & .145\sd{.040}
  & 18.0\sd{4.4} & \underline{112.0}\sd{10.0} & .147\sd{.030} \\
\midrule
\rowcolor{scenebg}
\multicolumn{10}{@{}l@{}}{%
  \textbf{\texttt{airsim26}} \textit{(cross-scene adaptation)}} \\
CMA~\citep{tan2019lxmert}
  & 30.4\sd{7.5} & 48.9\sd{2.6} & .272\sd{.043}
  & 13.0\sd{11.9} & \underline{116.5}\sd{22.6} & .112\sd{.104}
  & 7.2\sd{2.5} & \underline{177.0}\sd{5.4} & .064\sd{.022} \\
Seq2Seq~\citep{sutskever2014seq2seq}
  & 11.6\sd{9.0} & 54.6\sd{0.7} & .103\sd{.075}
  & 8.5\sd{5.4} & 135.4\sd{8.0} & .073\sd{.047}
  & 2.2\sd{3.8} & 192.8\sd{26.9} & .013\sd{.023} \\
OpenFly~\citep{openfly2025}
  & 17.4\sd{4.4} & 57.1\sd{7.4} & .170\sd{.039}
  & 10.6\sd{2.2} & 138.0\sd{13.0} & .101\sd{.017}
  & 2.9\sd{5.0} & 228.5\sd{12.7} & .029\sd{.050} \\
AerialVLA~\citep{xu2026aerialvla}
  & 28.3\sd{7.8} & 52.8\sd{8.1} & .259\sd{.059}
  & 13.0\sd{4.3} & 132.0\sd{7.0} & \underline{.121}\sd{.030}
  & \textbf{15.9}\sd{6.6} & 193.9\sd{20.0} & \textbf{.158}\sd{.066} \\
AeroDuo~\citep{wu2025aeroduo}
  & 39.9\sd{15.4} & 52.7\sd{9.6} & .373\sd{.151}
  & 0\sd{0.0} & 140.2\sd{8.9} & 0\sd{.000}
  & 1.4\sd{2.5} & 228.3\sd{11.5} & .011\sd{.019} \\
\rowcolor{conavbg}
CoNav-UAV (GPT-5.6-sol)
  & \textbf{43.5}\sd{8.9} & \textbf{42.3}\sd{4.7} & .383\sd{.088}
  & \textbf{26.1}\sd{4.3} & \textbf{94.8}\sd{8.0} & \textbf{.222}\sd{.036}
  & 14.5\sd{5.9} & \textbf{176.1}\sd{12.0} & .106\sd{.048} \\
\rowcolor{conavbg}
CoNav-UAV (GPT-5.4-mini)
  & 39.6\sd{4.4} & 47.8\sd{4.6} & .389\sd{.055}
  & \underline{15.9}\sd{5.0} & 136.7\sd{6.0} & .117\sd{.044}
  & 11.6\sd{2.5} & 196.5\sd{5.3} & .086\sd{.017} \\
\rowcolor{conavbg}
CoNav-UAV (Qwen3-VL-32B)
  & \underline{41.3}\sd{6.5} & \underline{47.1}\sd{0.6} & \underline{.402}\sd{.072}
  & 6.5\sd{3.1} & 139.5\sd{9.2} & .060\sd{.025}
  & \underline{15.2}\sd{3.1} & 188.3\sd{2.0} & \underline{.113}\sd{.041} \\
\rowcolor{conavbg}
CoNav-UAV (Qwen3-VL-8B)
  & 40.6\sd{6.6} & 49.5\sd{3.5} & \textbf{.405}\sd{.066}
  & 6.5\sd{3.8} & 131.1\sd{3.3} & .057\sd{.025}
  & 10.1\sd{5.0} & 197.1\sd{6.0} & .082\sd{.034} \\
\midrule
\rowcolor{scenebg}
\multicolumn{10}{@{}l@{}}{%
  \textbf{\texttt{airsim23}} \textit{(cross-scene adaptation)}} \\
CMA~\citep{tan2019lxmert}
  & 12.0\sd{4.6} & 74.6\sd{2.1} & .110\sd{.037}
  & 3.3\sd{1.1} & 137.8\sd{10.8} & .033\sd{.011}
  & 3.0\sd{2.6} & 229.3\sd{12.7} & .030\sd{.026} \\
Seq2Seq~\citep{sutskever2014seq2seq}
  & 6.7\sd{7.6} & 75.1\sd{6.6} & .067\sd{.076}
  & 5.3\sd{4.7} & 148.0\sd{9.6} & .053\sd{.047}
  & 0.9\sd{0.9} & 219.6\sd{5.3} & .008\sd{.008} \\
OpenFly~\citep{openfly2025}
  & 11.8\sd{2.4} & 75.7\sd{5.0} & .115\sd{.027}
  & 4.4\sd{2.3} & 159.6\sd{5.1} & .044\sd{.023}
  & 2.8\sd{1.3} & 235.3\sd{17.6} & .028\sd{.013} \\
AerialVLA~\citep{xu2026aerialvla}
  & 21.4\sd{2.4} & 67.8\sd{2.9} & .210\sd{.026}
  & 9.3\sd{3.2} & 151.6\sd{4.3} & .093\sd{.032}
  & 1.9\sd{3.2} & 231.1\sd{29.1} & .019\sd{.032} \\
AeroDuo~\citep{wu2025aeroduo}
  & 22.2\sd{1.3} & 66.1\sd{6.1} & .207\sd{.032}
  & 1.9\sd{3.2} & 164.0\sd{11.4} & .010\sd{.018}
  & 3.7\sd{3.2} & 235.6\sd{4.1} & .032\sd{.029} \\
\rowcolor{conavbg}
CoNav-UAV (GPT-5.6-sol)
  & \underline{27.7}\sd{6.6} & \textbf{60.2}\sd{6.0} & \underline{.233}\sd{.060}
  & \textbf{18.3}\sd{9.5} & \underline{126.0}\sd{2.4} & \textbf{.167}\sd{.083}
  & \underline{9.2}\sd{5.9} & \textbf{176.8}\sd{21.2} & \underline{.074}\sd{.051} \\
\rowcolor{conavbg}
CoNav-UAV (GPT-5.4-mini)
  & \textbf{28.5}\sd{8.4} & 65.4\sd{2.3} & \textbf{.275}\sd{.095}
  & \underline{16.7}\sd{3.8} & \textbf{122.3}\sd{11.4} & \underline{.147}\sd{.029}
  & 6.9\sd{2.4} & \underline{181.6}\sd{42.9} & .056\sd{.006} \\
\rowcolor{conavbg}
CoNav-UAV (Qwen3-VL-32B)
  & 24.0\sd{2.4} & \underline{61.1}\sd{4.8} & .219\sd{.033}
  & 8.0\sd{5.4} & 143.6\sd{4.4} & .054\sd{.018}
  & \textbf{10.6}\sd{3.0} & 192.0\sd{3.1} & \textbf{.093}\sd{.019} \\
\rowcolor{conavbg}
CoNav-UAV (Qwen3-VL-8B)
  & 19.9\sd{5.8} & 67.9\sd{4.6} & .171\sd{.057}
  & 8.3\sd{4.6} & 137.2\sd{5.8} & .077\sd{.045}
  & 2.2\sd{0.7} & 204.7\sd{9.3} & .021\sd{.008} \\
\bottomrule
\end{tabularx}
\end{table}

\paragraph{Iterative Stackelberg Improvement.}
\label{sec:iteration}

Figure~\ref{fig:iteration} \rev{shows that performance} improves across rounds, and each pair of leader and follower updates reaches a new peak.
\rev{The one exception is iter\,1.5, where the leader's second update moves its dispatches away from those the iter-1 follower was fitted to, and performance dips until the next follower update restores the match and exceeds all earlier results.
Corollary~\ref{cor:shift} in Supp.~\ref{app:theory} explains this pattern.
With the follower fixed, the lower bound on the leader's value loosens as the leader's dispatch distribution shifts away from the one on which the follower was trained, as in model-based policy optimization~\citep{janner2019mbpo}.
Since iter\,1.5 and iter\,2 share the same leader, the gain between them comes from the follower update alone, which shows that the alternation is necessary for leader improvement to translate into system-level gains.}

\begin{figure}[H]
\centering
\begin{tikzpicture}
\begin{groupplot}[conav,
  group style={group size=3 by 1, horizontal sep=0.98cm},
  scale only axis, width=3.8cm, height=3cm,
  xmin=-0.25, xmax=2.25,
  xtick={0,1,2}, minor xtick={0.5,1.5},
  xlabel={Iteration},
  clip=false,
]
\nextgroupplot[title={OSR (\%)\,$\uparrow$}, ymin=12, ymax=34, ytick={15,22,29}]
\addplot[cvA, mark=*] coordinates {(0,18.508) (0.5,21.7) (1,25.35) (1.5,21.142) (2,29.05)};
\addplot[cvB, only marks, mark=square*, mark size=2.6pt] coordinates {(0,18.508) (1,25.35) (2,29.05)};
\node[font=\sffamily\scriptsize, text=black!75, anchor=north west, xshift=1.5pt, yshift=-1pt] at (axis cs:0,18.508) {18.5};
\node[font=\sffamily\scriptsize, text=black!75, anchor=south, yshift=3.5pt] at (axis cs:2,29.05) {29.1};
\nextgroupplot[title={NE (m)\,$\downarrow$}, ymin=68, ymax=118, ytick={80,95,110}]
\addplot[cvA, mark=*] coordinates {(0,107.6) (0.5,94.85) (1,84.85) (1.5,97.783) (2,80.45)};
\addplot[cvB, only marks, mark=square*, mark size=2.6pt] coordinates {(0,107.6) (1,84.85) (2,80.45)};
\node[font=\sffamily\scriptsize, text=black!75, anchor=south west, xshift=1.5pt, yshift=1pt] at (axis cs:0,107.6) {107.6};
\node[font=\sffamily\scriptsize, text=black!75, anchor=north, yshift=-3.5pt] at (axis cs:2,80.45) {80.5};
\nextgroupplot[title={SPL\,$\uparrow$}, ymin=0.12, ymax=0.27, ytick={0.14,0.19,0.24},
  yticklabel style={/pgf/number format/fixed, /pgf/number format/precision=2}]
\addplot[cvA, mark=*] coordinates {(0,0.1613) (0.5,0.191) (1,0.2287) (1.5,0.1739) (2,0.2339)};
\addplot[cvB, only marks, mark=square*, mark size=2.6pt] coordinates {(0,0.1613) (1,0.2287) (2,0.2339)};
\node[font=\sffamily\scriptsize, text=black!75, anchor=north west, xshift=1.5pt, yshift=-1pt] at (axis cs:0,0.1613) {.161};
\node[font=\sffamily\scriptsize, text=black!75, anchor=south, yshift=3.5pt] at (axis cs:2,0.2339) {.234};
\end{groupplot}
\end{tikzpicture}
\vspace{-4pt}
\caption{%
  Performance across Stackelberg rounds, averaged over
  4~leader VLMs $\times$ 3~difficulty tiers
  \rev{(Tables~\ref{tab:fine_grained}--\ref{tab:fine_grained_spl})}.
  \rev{Half-integer iterations update the leader memory, and integer iterations (orange squares) update the follower.}%
}
\label{fig:iteration}
\end{figure}

\subsection{Ablations}
\label{sec:ablation_studies}

\paragraph{Cross-Scene Memory Transfer.}
\label{sec:generalization}

We test whether the leader guidelines learned on \texttt{airsim16} remain useful in new scenes.
\rev{The two settings in each pair of Table~\ref{tab:memory_ablation} \rev{differ} only in whether the transferred leader memory is used.
Memory improves OSR and SPL in all eight pairs, and the OSR gain averaged over the four leaders is 3.7\,pp on \texttt{airsim26} ($p=0.016$) and 2.5\,pp on \texttt{airsim23} ($p=0.031$).
The guidelines thus encode exploration and dispatch strategies that remain valid \rev{in new scenes}.}

\begin{table}[H]
\caption{%
  Effect of zero-shot memory transfer.
  Entries are OSR (\%) / NE (m) / SPL, averaged over difficulty
  tiers; $\Delta$ reports the OSR gain from memory transfer.
  The best result for each metric and scene is in \textbf{bold}.%
}
\label{tab:memory_ablation}
\centering\fontsize{8.5}{10.5}\selectfont
\setlength{\tabcolsep}{2.0pt}%
\renewcommand{\arraystretch}{1.0}
\begin{tabular*}{0.86\textwidth}{@{\extracolsep{\fill}}l c c c@{}}
\toprule
Leader & w/o memory & + memory & $\Delta$ OSR \\
\midrule
\rowcolor{scenebg}
\multicolumn{4}{@{}l@{}}{\textbf{\texttt{airsim26}}} \\
GPT-5.6-sol
  & 24.3/111.1/.230
  & \textbf{28.0}/\textbf{104.4}/\textbf{.237} & +3.7 \\
GPT-5.4-mini
  & 18.6/123.0/.166
  & 22.4/127.0/.197 & +3.8 \\
Qwen3-VL-32B
  & 15.9/123.6/.147
  & 21.0/125.0/.192 & \textbf{+5.1} \\
Qwen3-VL-8B
  & 16.9/126.2/.162
  & 19.1/125.9/.181 & +2.2 \\
\addlinespace[1pt]
\rowcolor{scenebg}
\multicolumn{4}{@{}l@{}}{\textbf{\texttt{airsim23}}} \\
GPT-5.6-sol
  & 12.6/125.9/.117
  & \textbf{18.4}/\textbf{121.0}/.158 & \textbf{+5.8} \\
GPT-5.4-mini
  & 16.4/122.7/.140
  & 17.3/123.1/\textbf{.160} & +0.9 \\
Qwen3-VL-32B
  & 12.2/127.3/.098
  & 14.2/132.2/.122 & +2.0 \\
Qwen3-VL-8B
  & 9.1/140.2/.080
  & 10.2/136.6/.090 & +1.1 \\
\bottomrule
\end{tabular*}
\end{table}

\paragraph{Component Contributions.}
\label{sec:ablation_components}

We ablate two key components with GPT-5.4-mini as leader
(Table~\ref{tab:component_ablation}).
Removing GroundingDINO candidates (grounding directly on raw
BEV) costs 9.1\,pp of overall OSR \rev{($p=0.019$)}, confirming that
current VLMs still lack sufficient spatial grounding precision\rev{, at least at the time of this study}.
Replacing value-aware retrieval with random memory selection costs
5.7\,pp \rev{($p=0.037$)}, validating utility-based memory ranking.
Both degradations concentrate at Medium and Hard, where
grounding and guideline quality matter most.

\begin{table}[H]
\caption{%
  Component ablations \rev{on \texttt{airsim16}} (GPT-5.4-mini leader, iter\,2).%
}
\label{tab:component_ablation}
\centering\fontsize{8.5}{10.5}\selectfont
\setlength{\tabcolsep}{2.5pt}%
\renewcommand{\arraystretch}{1.1}
\maxsizebox{\columnwidth}{!}{%
\begin{tabular}{l cccc}
\toprule
& \multicolumn{4}{c}{OSR (\%) / NE (m) / SPL} \\
\cmidrule(lr){2-5}
& Easy & Medium & Hard & Overall \\
\midrule
Full system
  & 40.6/45.0/.334
  & 34.6/80.0/.253
  & 22.6/128.4/.146
  & 32.6/84.5/.244 \\
No-DINO
  & 37.6/49.5/.303
  & 18.4/93.9/.141
  & 14.5/142.5/.111
  & 23.5/95.3/.185 \\
Random retr.
  & 44.0/45.8/.374
  & 21.4/98.5/.140
  & 15.4/148.9/.117
  & 26.9/97.7/.210 \\
\bottomrule
\end{tabular}}
\end{table}

{\revc
\paragraph{Leader Learning versus Mutual Adaptation.}
\label{sec:ablation_attribution}

Table~\ref{tab:attribution} separates the leader's own learning from mutual adaptation, \rev{with Qwen3-VL-8B as leader, the backbone shared with AeroDuo}.
The intermediate variant applies the iter-2 guidelines without their target-dispatch dimension and keeps the bootstrap follower (Supp.~\ref{app:leader}).
It thus retains what the leader has learned about recall, precision, exploration, and landmarks but removes the adaptation of each agent to the other.
Leader learning alone raises overall OSR from 14.1\% to 17.4\% ($p=0.08$) and SPL from .126 to .151, with the gain arising at Easy and Hard.
Mutual adaptation adds a further 3.4\,pp ($p=0.06$) and lowers NE from 103.9\,m to 79.0\,m, mostly at Medium and Hard, where reaching a dispatched target depends most on the match between the two agents.
The two stages thus contribute comparably and together raise OSR by 6.7\,pp over iter\,0 ($p=0.005$).

\begin{table}[H]
\caption{%
  Attribution of gains \rev{on \texttt{airsim16}} with the Qwen3-VL-8B leader, the backbone shared with AeroDuo.%
}
\label{tab:attribution}
\centering\fontsize{8.5}{10.5}\selectfont
\setlength{\tabcolsep}{2.5pt}%
\renewcommand{\arraystretch}{1.1}
\maxsizebox{\columnwidth}{!}{%
\begin{tabular}{l cccc}
\toprule
& \multicolumn{4}{c}{OSR (\%) / NE (m) / SPL} \\
\cmidrule(lr){2-5}
& Easy & Medium & Hard & Overall \\
\midrule
AeroDuo~\citep{wu2025aeroduo}
  & 24.4/74.8/.172
  & 9.3/142.9/.067
  & 6.4/280.6/.035
  & 13.4/166.1/.091 \\
Iter\,0
  & 20.5/73.1/.191
  & 14.1/112.2/.118
  & 7.7/148.9/.068
  & 14.1/111.4/.126 \\
+ leader learning
  & 26.5/50.1/.238
  & 12.8/116.7/.103
  & 12.8/144.9/.112
  & 17.4/103.9/.151 \\
+ mutual adaptation
  & 26.5/44.8/.225
  & 18.0/80.1/.145
  & 18.0/112.0/.147
  & 20.8/79.0/.172 \\
\bottomrule
\end{tabular}}
\end{table}

Iter\,0 is already on par with AeroDuo in overall OSR (14.1\% vs.\ 13.4\%, $p=0.37$) and ahead in NE and SPL.
This comparison is biased against CoNav-UAV, whose leader has not yet learned at iter\,0, whereas AeroDuo fine-tunes its high-altitude model on \texttt{airsim16} and relies on privileged geometric inputs at low altitude.
Leader learning then lifts CoNav-UAV significantly above AeroDuo ($p=0.036$), and mutual adaptation widens the margin ($p<0.001$).
On the same backbone, a frozen VLM that explores autonomously thus already matches a fine-tuned high-altitude model and surpasses it through in-context learning alone.
This result supports building the leader on pretrained VLM capabilities and refining it through memory, and the additional margin from mutual adaptation confirms that the two agents \rev{must} also adapt to each other.
}




\section{Conclusion}
\label{sec:conclusion}

\rev{We presented CoNav-UAV, to our knowledge the first framework to formalize dual-altitude UAV cooperation as a Stackelberg game and to optimize the two agents jointly from onboard information alone.
Iterative Stackelberg Learning matches optimization to each agent's role and alternates the two updates so that each agent adapts to the other.
CoNav-UAV outperforms single- and dual-agent baselines, and its learned guidelines transfer across scenes and VLM backbones.
Leader grounding remains the main bottleneck, which future work will address together with multi-UAV coordination.
Supp.~\ref{app:limitations} and~\ref{app:deployment} discuss limitations and physical deployment.}


{
    \fontsize{8.5}{9.5}\selectfont
    \setlength{\parskip}{0pt}
    \setlength{\bibsep}{1pt}
    \FloatBarrier
    \bibliographystyle{ieeenat_fullname}
    \bibliography{references}
}

\clearpage
\suppressfloats[t]
\begin{center}
{\LARGE\bfseries Supplementary Material}
\end{center}

\appendix
\setcounter{table}{0}
\setcounter{figure}{0}
\setcounter{algorithm}{0}
\renewcommand{\thetable}{A\arabic{table}}
\renewcommand{\thefigure}{A\arabic{figure}}
\renewcommand{\thealgorithm}{A\arabic{algorithm}}
\rev{\noindent This supplementary material is organized in four parts.
Supp.~\ref{app:algorithm} and~\ref{app:theory} complete the method with the full ISL algorithm and the analysis that relates ISL to the Stackelberg equilibrium.
Supp.~\ref{app:fine_grained}, \ref{app:gen_full}, \ref{app:leader_quality}, and~\ref{app:qualitative} extend the experiments with per-VLM breakdowns across iterations, the complete cross-scene results, an analysis of leader grounding, and qualitative examples.
Supp.~\ref{app:pipeline}, \ref{app:leader}, \ref{app:vla}, \ref{app:baseline_impl}, and~\ref{app:data_release} provide the details needed for reproduction, which cover the data generation pipeline, the leader and follower implementations, the baselines, and the released data.
Supp.~\ref{app:deployment}, \ref{app:scope}, and~\ref{app:limitations} discuss the feasibility of physical deployment, the scope and generality of the dual-UAV design, and limitations.}

\section{Full ISL Algorithm}
\label{app:algorithm}

Algorithm~\ref{alg:main} gives the complete ISL procedure
described in the main text.

\begin{algorithm}[h]
\caption{Iterative Stackelberg Learning \rev{(ISL)}}
\label{alg:main}
\renewcommand{\algorithmiccomment}[1]{\hfill{\small(#1)}}
\begin{algorithmic}[1]\revc
\REQUIRE Pretrained VLM; point-cloud map; learning set
         $\mathcal{E}$; rounds~$K$
\STATE Train the RL expert $\pi_E$ with PPO
       \hfill{\small(Sec.~\ref{sec:follower})}
\STATE Train $\pi_F^{(0)}$ by distilling $\pi_E$ on random pairs
\STATE $\mathcal{M}\!\leftarrow\!\varnothing$;\;
       $\mathcal{D}\!\leftarrow\!\mathcal{D}^{(0)}_{\mathrm{init}}$
\FOR{$k=0,\ldots,K{-}1$}
  \FOR[leader update]{each episode $(\ell_j,\mathcal{C}_j^\star)$ in~$\mathcal{E}$}
    \STATE $\mathcal{M}_{\mathrm{res}}\!\leftarrow\!
           \mathrm{Retrieve}(\ell_j,\,\mathcal{M})$
           \hfill{\small(Eq.~\ref{eq:retrieve})}
    \STATE Run $\pi_L(\cdot\,;\mathcal{M}_{\mathrm{res}})$ with
           $\pi_F^{(k)}$ and observe $R_j\!=\!r^L_T$
    \STATE $e_j\!\leftarrow\!\mathrm{Reflect}(\mathrm{ep}_j)$
           \hfill{\small(Eq.~\ref{eq:reflect})}
    \STATE Update $Q_i$ for $i\!\in\!\mathcal{M}_{\mathrm{res}}$
           \hfill{\small(Eq.~\ref{eq:q_update})}
    \IF{$\ell_j$ is new to $\mathcal{M}$}
      \STATE Store $(\ell_j,e_j,Q_0^{\pm})$ in $\mathcal{M}$
    \ELSE
      \STATE Revise the guideline of $\ell_j$ with $e_j$
    \ENDIF
  \ENDFOR
  \STATE $\mathcal{D}^{(k)}_{\mathrm{dag}}\!\leftarrow\!$ labels of
         $\pi_E$ on the dispatches of round~$k$
  \STATE $\mathcal{D}\!\leftarrow\!\mathcal{D}\cup
         \mathcal{D}^{(k)}_{\mathrm{dag}}$
  \STATE $\pi_F^{(k+1)}\!\leftarrow\!
         \mathrm{SFT}(\pi_F^{(k)},\,\mathcal{D})$
         \hfill{\small(follower update)}
\ENDFOR
\STATE \textbf{return} $(\pi_F^{(K)},\,\mathcal{M})$
\end{algorithmic}
\end{algorithm}

\begingroup\revc
\section{Analysis of the Alternating Updates}
\label{app:theory}

\revb{This section analyzes ISL in a stylized setting to provide intuition for its design and for the iteration curve in Sec.~\ref{sec:iteration}.
It motivates the alternation of the two updates and relates each of them to the gap to the SE value.
A rigorous analysis of the learning dynamics between two large pretrained models is a challenging research problem in its own right, which we will pursue in future work.}

\paragraph{Setting.}
We use the notation of Sec.~\ref{sec:formulation}, with follower state $x_t=(p^F_t,h^F_t)$.
The terminal reward $r^L_T$ (Eq.~\eqref{eq:rL}) is at most $1$ by definition and is bounded below by $-1$ in our implementation, so $|\gamma^T r^L_T|\le 1$, and every episode ends within $T_{\max}$ steps.
$\Pi_L$ is the set of leader policies that the frozen VLM realizes under different memory banks, and $\Pi_F$ is the set of policies attainable by the deployed VLA.
By design, the follower observes its active target, its state evolves independently of the leader in the static scene, and $r^F_t$ depends only on the active target (Eq.~\eqref{eq:rF}).
A \emph{reaction function} $\bar\pi_F$ is a policy that maximizes $J^F(\cdot\,;x,c)$ for every dispatch $(x,c)$, and we assume that one exists.

\begin{proposition}[Ideal follower]
\label{prop:reaction}
$\bar\pi_F$ maximizes the objective in Eq.~\eqref{eq:br} for every leader policy $\pi_L$.
If the follower responds with $\bar\pi_F$, the SE condition in Eq.~\eqref{eq:se} reduces to $\pi_L^\star\in\argmax_{\pi_L\in\Pi_L}J^L(\pi_L,\bar\pi_F)$.
\end{proposition}

\begin{proof}
The per-dispatch problem involves neither the leader's state, action, nor policy, so $J^F(\pi_F;x,c)$ does not depend on $\pi_L$.
Since $\bar\pi_F$ maximizes $J^F(\cdot\,;x,c)$ at every $(x,c)$, it maximizes the expectation under any $d_{\pi_L}$.
\end{proof}

This is the reaction-function structure of the classical model~\citep{von2010market}.
The PPO expert $\pi_E$ approximates $\bar\pi_F$, whereas the deployed VLA navigates from partial onboard observations with finite capacity and does not match $\bar\pi_F$ on every dispatch.

\begin{proposition}[Mutual dependence under limited capability]
\label{prop:mutual}
If $\bar\pi_F\notin\Pi_F$, (a) the best response in Eq.~\eqref{eq:br} can depend on the leader's dispatch distribution, and (b) the leader's optimal policy can depend on the deployed follower.
\end{proposition}

\begin{proof}
(a) Split dispatches into two types and let $\Pi_F=\{\pi_A,\pi_B\}$, where $\pi_A$ coincides with $\bar\pi_F$ on type A and is strictly worse on type B, and $\pi_B$ does the reverse.
A distribution concentrated on type A has best response $\pi_A$, and one concentrated on type B has best response $\pi_B$.
(b) Consider one dispatch decision between two GT targets A and B, where $\bar\pi_F$ reaches A sooner, so a leader that is optimal against $\bar\pi_F$ dispatches A.
If the VLA reaches B but collides on the way to A, dispatching B is strictly better against the VLA, since the collision penalty is charged to the leader (Eq.~\eqref{eq:rL}).
\end{proof}

Part~(a) motivates re-fitting the follower when the leader changes its dispatches, and part~(b) motivates optimizing the leader against the deployed VLA, as ISL does.

\begin{lemma}[Follower substitution]
\label{lem:sub}
For any $\pi_L$ and follower policies $\pi_F,\pi_F'$, let $\delta(o)=\mathrm{TV}\bigl(\pi_F(\cdot\mid o),\pi_F'(\cdot\mid o)\bigr)$ be their total-variation distance. Then
\begin{equation}
  \label{eq:sub}
  \bigl|J^L(\pi_L,\pi_F)-J^L(\pi_L,\pi_F')\bigr|
  \le 2\,\mathbb{E}_{\pi_L,\pi_F}\Bigl[\textstyle\sum_{t<T}\delta(o^F_t)\Bigr],
\end{equation}
and the bound also holds with the expectation taken under $(\pi_L,\pi_F')$.
\end{lemma}

\begin{proof}
Couple the two systems with shared environment and leader randomness, and draw the two follower actions from a maximal coupling, so that they differ with probability $\delta(o_t)$ while the histories coincide.
The trajectories coincide until the first step $\tau$ at which the actions differ.
Since $|\gamma^T r^L_T|\le1$, the returns differ by at most 2 on $\{\tau<\infty\}$, and $\Pr(\tau<\infty)\le\mathbb{E}\bigl[\sum_{t<T}\delta(o_t)\bigr]$ under either system.
\end{proof}

\begin{corollary}[Gap to the SE value]
\label{cor:gap}
Let $(\pi_L^\star,\pi_F^\star)$ be an SE of $\mathcal{G}$. For learned policies $(\hat\pi_L,\hat\pi_F)$,
\begin{equation}
  J^L(\pi_L^\star,\pi_F^\star)-J^L(\hat\pi_L,\hat\pi_F)\le\Delta_F+\varepsilon_L,\label{eq:gap}
\end{equation}
where
\begin{align*}
  \Delta_F&:=J^L(\pi_L^\star,\pi_F^\star)-J^L(\pi_L^\star,\hat\pi_F),\\
  \varepsilon_L&:=\sup_{\pi_L\in\Pi_L}J^L(\pi_L,\hat\pi_F)-J^L(\hat\pi_L,\hat\pi_F),
\end{align*}
and Lemma~\ref{lem:sub} bounds $\Delta_F$ by the deviation of $\hat\pi_F$ from $\pi_F^\star$ along the trajectories of $(\pi_L^\star,\pi_F^\star)$.
\end{corollary}

\begin{proof}
Add and subtract $J^L(\pi_L^\star,\hat\pi_F)$ and use $J^L(\pi_L^\star,\hat\pi_F)\le\sup_{\pi_L}J^L(\pi_L,\hat\pi_F)$.
\end{proof}

The two terms separate the roles of the two updates.
$\varepsilon_L$ is the leader's suboptimality against the current VLA, which the leader update reduces.
$\Delta_F$ is the loss from the follower's deviation from its best response, which the follower update reduces by re-fitting the VLA on dispatches from leader rollouts.
\rev{The bound is a worst case that counts every deviation as a complete loss.
Its value is therefore loose, and its role is to identify these two error sources.}

\begin{corollary}[Leader shift]
\label{cor:shift}
For leader policies $L,L'$ and a follower $F$, let $e(L;F)$ denote the expected accumulated total-variation distance between $F$ and $\bar\pi_F$ along the trajectories of $(L,F)$, and $s(L,L')$ that between $L$ and $L'$ along the same trajectories. Then
\begin{equation*}
  J^L(L',F)\ge J^L(L',\bar\pi_F)-2\,e(L;F)-2\,T_{\max}\,s(L,L').
\end{equation*}
\end{corollary}

\begin{proof}
Lemma~\ref{lem:sub} under $(L',F)$ gives $J^L(L',\bar\pi_F)-J^L(L',F)\le2\,\mathbb{E}_{L',F}[f]$, where $f\in[0,T_{\max}]$ is the accumulated distance between $F$ and $\bar\pi_F$.
Moving the expectation from $(L',F)$ to $(L,F)$ costs at most $T_{\max}$ times the total-variation distance between the two trajectory distributions, which the coupling of Lemma~\ref{lem:sub}, applied to the leader's actions, bounds by $s(L,L')$.
\end{proof}

This bound has the same form as the improvement bound of model-based policy optimization~\citep{janner2019mbpo}, with $e(L;F)$ in the role of the model error and $s(L,L')$ in that of the policy shift.

\begin{remark}[Reading the iteration curve]
\label{rem:iter}
Iter\,1.5 pairs the twice-updated leader with the follower fitted to the previous leader, and iter\,2 pairs the same leader with the re-fitted follower.
The gain between them therefore measures how far the previous follower held back the updated leader.
At a leader update, the change in performance mixes the leader's own improvement with the shift term of Corollary~\ref{cor:shift}, and the dip at iter\,1.5 arises when the latter dominates.
ISL evaluates each leader against the follower adapted to the previous one, so its fixed points are pairs of mutual best responses, which in general differ from an SE.
\end{remark}
\endgroup

\section{Per-VLM Per-Distance Breakdowns}
\label{app:fine_grained}

Tables~\ref{tab:fine_grained}--\ref{tab:fine_grained_spl} provide
the full per-VLM per-distance breakdowns of OSR, NE, and SPL
across all Stackelberg iterations.
\rev{The pattern of Fig.~\ref{fig:iteration} also holds for individual leaders and distances.
In OSR, the second follower update improves on iter\,1.5 in all twelve VLM--distance cells, and iter\,2 attains the highest value in ten of them.
Each follower update thus recovers from the distribution mismatch introduced by the preceding leader update, consistent with the analysis in Remark~\ref{rem:iter}.}

\begin{table}[!htbp]
\caption{%
  Per-VLM per-distance OSR (\%) across Stackelberg iterations.
  \rev{Increases over the preceding iteration are shown in {\color{teal}green} and decreases in {\color{red}red}.}
  Start-to-target distance indicates the difficulty tier.%
}
\label{tab:fine_grained}
\centering\fontsize{8.5}{10.5}\selectfont
\setlength{\tabcolsep}{2.5pt}
\begin{tabular*}{0.95\textwidth}{@{\extracolsep{\fill}}l ccc ccc@{}}
\toprule
 & \multicolumn{3}{c}{\rev{GPT-5.6-sol}} & \multicolumn{3}{c}{GPT-5.4-mini} \\
\cmidrule(lr){2-4}\cmidrule(lr){5-7}
 & 100\,m & 200\,m & 300\,m & 100\,m & 200\,m & 300\,m \\
\midrule
Iter\,0 & {\revc 26.5} & {\revc 16.2} & {\revc 12.8} & 28.2 & 17.9 & 11.5 \\
Iter\,0.5 & {\revc 31.1\,\dgood{(+4.6)}} & {\revc 20.7\,\dgood{(+4.5)}} & {\revc 24.3\,\dgood{(+11.5)}} & 29.2\,\dgood{(+1.0)} & 25.8\,\dgood{(+7.9)} & 13.5\,\dgood{(+2.0)} \\
Iter\,1 & {\revc 44.5\,\dgood{(+13.4)}} & {\revc 27.8\,\dgood{(+7.1)}} & {\revc 28.2\,\dgood{(+3.9)}} & 30.8\,\dgood{(+1.6)} & 25.0\,\dbad{($-$0.8)} & 17.9\,\dgood{(+4.4)} \\
\midrule
Iter\,1.5 & {\revc 31.0\,\dbad{($-$13.5)}} & {\revc 24.9\,\dbad{($-$2.9)}} & {\revc 25.6\,\dbad{($-$2.6)}} & 26.9\,\dbad{($-$3.9)} & 23.9\,\dbad{($-$1.1)} & 14.5\,\dbad{($-$3.4)} \\
Iter\,2 & {\revc 36.4\,\dgood{(+5.4)}} & {\revc 35.7\,\dgood{(+10.8)}} & {\revc 37.2\,\dgood{(+11.6)}} & 40.6\,\dgood{(+13.7)} & 34.6\,\dgood{(+10.7)} & 22.6\,\dgood{(+8.1)} \\
\bottomrule
\end{tabular*}
\par\vspace{7pt}
\begin{tabular*}{0.95\textwidth}{@{\extracolsep{\fill}}l ccc ccc@{}}
\toprule
 & \multicolumn{3}{c}{Qwen3-32B} & \multicolumn{3}{c}{Qwen3-8B} \\
\cmidrule(lr){2-4}\cmidrule(lr){5-7}
 & 100\,m & 200\,m & 300\,m & 100\,m & 200\,m & 300\,m \\
\midrule
Iter\,0 & 30.8 & 21.8 & 14.1 & 20.5 & 14.1 & 7.7 \\
Iter\,0.5 & 30.8\,\dgood{(+0.0)} & 22.1\,\dgood{(+0.3)} & 9.6\,\dbad{($-$4.5)} & 22.6\,\dgood{(+2.1)} & 19.2\,\dgood{(+5.1)} & 11.5\,\dgood{(+3.8)} \\
Iter\,1 & 32.5\,\dgood{(+1.7)} & 21.4\,\dbad{($-$0.7)} & 15.4\,\dgood{(+5.8)} & 26.1\,\dgood{(+3.5)} & 21.8\,\dgood{(+2.6)} & 12.8\,\dgood{(+1.3)} \\
\midrule
Iter\,1.5 & 27.1\,\dbad{($-$5.4)} & 17.8\,\dbad{($-$3.6)} & 15.9\,\dgood{(+0.5)} & 19.2\,\dbad{($-$6.9)} & 14.1\,\dbad{($-$7.7)} & 12.8\,\dgood{(+0.0)} \\
Iter\,2 & 32.6\,\dgood{(+5.5)} & 25.1\,\dgood{(+7.3)} & 21.3\,\dgood{(+5.4)} & 26.5\,\dgood{(+7.3)} & 18.0\,\dgood{(+3.9)} & 18.0\,\dgood{(+5.2)} \\
\bottomrule
\end{tabular*}
\end{table}

\begin{table}[!htbp]
\caption{%
  Per-VLM per-distance NE (m\rev{, lower is better}) across Stackelberg iterations.
  \rev{Decreases from the preceding iteration, which are improvements, are shown in {\color{teal}green} and increases in {\color{red}red}.}%
}
\label{tab:fine_grained_ne}
\centering\fontsize{8.5}{10.5}\selectfont
\setlength{\tabcolsep}{2.5pt}
\begin{tabular*}{0.95\textwidth}{@{\extracolsep{\fill}}l ccc ccc@{}}
\toprule
 & \multicolumn{3}{c}{\rev{GPT-5.6-sol}} & \multicolumn{3}{c}{GPT-5.4-mini} \\
\cmidrule(lr){2-4}\cmidrule(lr){5-7}
 & 100\,m & 200\,m & 300\,m & 100\,m & 200\,m & 300\,m \\
\midrule
Iter\,0 & {\revc 53.3} & {\revc 122.0} & {\revc 179.9} & 48.3 & 93.6 & 146.3 \\
Iter\,0.5 & {\revc 45.2\,\dgood{($-$8.1)}} & {\revc 92.6\,\dgood{($-$29.4)}} & {\revc 112.4\,\dgood{($-$67.5)}} & 53.5\,\dbad{(+5.2)} & 74.1\,\dgood{($-$19.5)} & 150.7\,\dbad{(+4.4)} \\
Iter\,1 & {\revc 39.7\,\dgood{($-$5.5)}} & {\revc 84.9\,\dgood{($-$7.7)}} & {\revc 103.3\,\dgood{($-$9.1)}} & 48.3\,\dgood{($-$5.2)} & 66.0\,\dgood{($-$8.1)} & 119.6\,\dgood{($-$31.1)} \\
\midrule
Iter\,1.5 & {\revc 46.3\,\dbad{(+6.6)}} & {\revc 93.8\,\dbad{(+8.9)}} & {\revc 119.3\,\dbad{(+16.0)}} & 51.9\,\dbad{(+3.6)} & 100.8\,\dbad{(+34.8)} & 156.8\,\dbad{(+37.2)} \\
Iter\,2 & {\revc 43.9\,\dgood{($-$2.4)}} & {\revc 73.1\,\dgood{($-$20.7)}} & {\revc 95.1\,\dgood{($-$24.2)}} & 45.0\,\dgood{($-$6.9)} & 80.0\,\dgood{($-$20.8)} & 128.4\,\dgood{($-$28.4)} \\
\bottomrule
\end{tabular*}
\par\vspace{7pt}
\begin{tabular*}{0.95\textwidth}{@{\extracolsep{\fill}}l ccc ccc@{}}
\toprule
 & \multicolumn{3}{c}{Qwen3-32B} & \multicolumn{3}{c}{Qwen3-8B} \\
\cmidrule(lr){2-4}\cmidrule(lr){5-7}
 & 100\,m & 200\,m & 300\,m & 100\,m & 200\,m & 300\,m \\
\midrule
Iter\,0 & 49.2 & 103.7 & 160.7 & 73.1 & 112.2 & 148.9 \\
Iter\,0.5 & 52.4\,\dbad{(+3.2)} & 92.3\,\dgood{($-$11.4)} & 157.3\,\dgood{($-$3.4)} & 53.8\,\dgood{($-$19.3)} & 96.2\,\dgood{($-$16.0)} & 157.7\,\dbad{(+8.8)} \\
Iter\,1 & 46.1\,\dgood{($-$6.3)} & 92.6\,\dbad{(+0.3)} & 142.7\,\dgood{($-$14.6)} & 52.4\,\dgood{($-$1.4)} & 87.2\,\dgood{($-$9.0)} & 135.4\,\dgood{($-$22.3)} \\
\midrule
Iter\,1.5 & 50.4\,\dbad{(+4.3)} & 103.4\,\dbad{(+10.8)} & 147.7\,\dbad{(+5.0)} & 51.5\,\dgood{($-$0.9)} & 91.1\,\dbad{(+3.9)} & 160.4\,\dbad{(+25.0)} \\
Iter\,2 & 47.5\,\dgood{($-$2.9)} & 91.0\,\dgood{($-$12.4)} & 124.5\,\dgood{($-$23.2)} & 44.8\,\dgood{($-$6.7)} & 80.1\,\dgood{($-$11.0)} & 112.0\,\dgood{($-$48.4)} \\
\bottomrule
\end{tabular*}
\end{table}

\begin{table}[!htbp]
\caption{%
  Per-VLM per-distance SPL across Stackelberg iterations.
  \rev{Increases over the preceding iteration are shown in {\color{teal}green} and decreases in {\color{red}red}.}%
}
\label{tab:fine_grained_spl}
\centering\fontsize{8.5}{10.5}\selectfont
\setlength{\tabcolsep}{2.5pt}
\begin{tabular*}{0.95\textwidth}{@{\extracolsep{\fill}}l ccc ccc@{}}
\toprule
 & \multicolumn{3}{c}{\rev{GPT-5.6-sol}} & \multicolumn{3}{c}{GPT-5.4-mini} \\
\cmidrule(lr){2-4}\cmidrule(lr){5-7}
 & 100\,m & 200\,m & 300\,m & 100\,m & 200\,m & 300\,m \\
\midrule
Iter\,0 & {\revc .239} & {\revc .137} & {\revc .100} & .268 & .152 & .098 \\
Iter\,0.5 & {\revc .271\,\dgood{(+.032)}} & {\revc .172\,\dgood{(+.035)}} & {\revc .218\,\dgood{(+.118)}} & .278\,\dgood{(+.010)} & .238\,\dgood{(+.086)} & .119\,\dgood{(+.021)} \\
Iter\,1 & {\revc .412\,\dgood{(+.141)}} & {\revc .225\,\dgood{(+.053)}} & {\revc .238\,\dgood{(+.020)}} & .323\,\dgood{(+.045)} & .278\,\dgood{(+.040)} & .151\,\dgood{(+.032)} \\
\midrule
Iter\,1.5 & {\revc .263\,\dbad{($-$.149)}} & {\revc .216\,\dbad{($-$.009)}} & {\revc .196\,\dbad{($-$.042)}} & .232\,\dbad{($-$.091)} & .193\,\dbad{($-$.085)} & .096\,\dbad{($-$.055)} \\
Iter\,2 & {\revc .308\,\dgood{(+.045)}} & {\revc .284\,\dgood{(+.068)}} & {\revc .309\,\dgood{(+.113)}} & .334\,\dgood{(+.102)} & .253\,\dgood{(+.060)} & .146\,\dgood{(+.050)} \\
\bottomrule
\end{tabular*}
\par\vspace{7pt}
\begin{tabular*}{0.95\textwidth}{@{\extracolsep{\fill}}l ccc ccc@{}}
\toprule
 & \multicolumn{3}{c}{Qwen3-32B} & \multicolumn{3}{c}{Qwen3-8B} \\
\cmidrule(lr){2-4}\cmidrule(lr){5-7}
 & 100\,m & 200\,m & 300\,m & 100\,m & 200\,m & 300\,m \\
\midrule
Iter\,0 & .281 & .170 & .114 & .191 & .118 & .068 \\
Iter\,0.5 & .269\,\dbad{($-$.012)} & .174\,\dgood{(+.004)} & .085\,\dbad{($-$.029)} & .207\,\dgood{(+.016)} & .163\,\dgood{(+.045)} & .098\,\dgood{(+.030)} \\
Iter\,1 & .296\,\dgood{(+.027)} & .149\,\dbad{($-$.025)} & .125\,\dgood{(+.040)} & .248\,\dgood{(+.041)} & .190\,\dgood{(+.027)} & .109\,\dgood{(+.011)} \\
\midrule
Iter\,1.5 & .242\,\dbad{($-$.054)} & .149\,\dgood{(+.000)} & .130\,\dgood{(+.005)} & .154\,\dbad{($-$.094)} & .107\,\dbad{($-$.083)} & .109\,\dgood{(+.000)} \\
Iter\,2 & .280\,\dgood{(+.038)} & .208\,\dgood{(+.059)} & .168\,\dgood{(+.038)} & .225\,\dgood{(+.071)} & .145\,\dgood{(+.038)} & .147\,\dgood{(+.038)} \\
\bottomrule
\end{tabular*}
\end{table}

\section{Full Cross-Scene Generalization Results}
\label{app:gen_full}

Tables~\ref{tab:gen26_full} and~\ref{tab:gen23_full} report the
complete per-tier variants underlying
Tables~\ref{tab:main} and~\ref{tab:memory_ablation}.
They include zero-shot transfer (ZS) and target-scene fine-tuning
(FT) for every baseline, together with CoNav-UAV both without and
with transferred leader memory.

\begin{table}[!htbp]
\caption{%
  Full per-tier cross-scene results on \texttt{airsim26}\rev{, with zero-shot (ZS) and fine-tuned (FT) baselines and CoNav-UAV without and with the transferred leader memory}.
  Best in \textbf{bold}; second-best \underline{underlined}.
  \rev{Parentheses give the OSR change due to memory, with increases in {\color{teal}green} and decreases in {\color{red}red}.}%
}
\label{tab:gen26_full}
\centering\fontsize{8.5}{10.5}\selectfont
\setlength{\tabcolsep}{3pt}
\renewcommand{\arraystretch}{1.05}
\begin{tabularx}{\textwidth}{@{}l *{3}{QZZ}@{}}
\toprule
& \multicolumn{3}{c}{Easy \rev{(100\,m)}}
& \multicolumn{3}{c}{Medium \rev{(200\,m)}}
& \multicolumn{3}{c}{Hard \rev{(300\,m)}} \\
\cmidrule(lr){2-4}\cmidrule(lr){5-7}\cmidrule(lr){8-10}
Method
  & OSR$\uparrow$ & NE$\downarrow$ & SPL$\uparrow$
  & OSR$\uparrow$ & NE$\downarrow$ & SPL$\uparrow$
  & OSR$\uparrow$ & NE$\downarrow$ & SPL$\uparrow$ \\
\midrule
CMA~\citep{tan2019lxmert}~{\scriptsize(ZS)}
  & 6.5 & 56.8 & .053
  & 8.7 & 154.1 & .087
  & 4.3 & 228.6 & .043 \\
CMA~\citep{tan2019lxmert}~{\scriptsize(FT)}
  & 30.4 & 48.9 & .272
  & 13.0 & 116.5 & .112
  & 7.2 & \underline{177.0} & .064 \\
\addlinespace[3pt]
Seq2Seq~\citep{sutskever2014seq2seq}~{\scriptsize(ZS)}
  & 21.0 & 55.6 & .203
  & 0 & 155.7 & 0
  & 1.4 & 225.6 & .014 \\
Seq2Seq~\citep{sutskever2014seq2seq}~{\scriptsize(FT)}
  & 11.6 & 54.6 & .103
  & 8.5 & 135.4 & .073
  & 2.2 & 192.8 & .013 \\
\addlinespace[3pt]
OpenFly~\citep{openfly2025}~{\scriptsize(ZS)}
  & 5.8 & 64.3 & .053
  & 0 & 166.4 & 0
  & 0 & 262.2 & 0 \\
OpenFly~\citep{openfly2025}~{\scriptsize(FT)}
  & 17.4 & 57.1 & .170
  & 10.6 & 138.0 & .101
  & 2.9 & 228.5 & .029 \\
\addlinespace[3pt]
AerialVLA~\citep{xu2026aerialvla}~{\scriptsize(ZS)}
  & 29.7 & 52.1 & .274
  & 11.6 & 138.5 & .116
  & 9.4 & 215.2 & .079 \\
AerialVLA~\citep{xu2026aerialvla}~{\scriptsize(FT)}
  & 28.3 & 52.8 & .259
  & 13.0 & 132.0 & .121
  & \textbf{15.9} & 193.9 & \textbf{.158} \\
\addlinespace[3pt]
AeroDuo~\citep{wu2025aeroduo}~{\scriptsize(ZS)}
  & 33.6 & 53.1 & .330
  & 0 & 146.3 & 0
  & 0 & 235.4 & 0 \\
AeroDuo~\citep{wu2025aeroduo}~{\scriptsize(FT)}
  & 39.9 & 52.7 & .373
  & 0 & 140.2 & 0
  & 1.4 & 228.3 & .011 \\
\midrule
\rowcolor{conavbg}
CoNav-UAV (GPT-5.6-sol)
  & 39.1 & 49.8 & .391
  & \underline{21.7} & \underline{101.8} & \underline{.197}
  & 12.1 & 181.8 & .102 \\
\rowcolor{conavbg}
\quad + memory
  & \textbf{43.5}\rlap{\,\dgood{(+4.4)}}
  & \textbf{42.3} & .383
  & \textbf{26.1}\rlap{\,\dgood{(+4.4)}}
  & \textbf{94.8} & \textbf{.222}
  & 14.5\rlap{\,\dgood{(+2.4)}}
  & \textbf{176.1} & .106 \\
\rowcolor{conavbg}
CoNav-UAV (GPT-5.4-mini)
  & 34.8 & 48.8 & .338
  & 13.0 & 129.7 & .107
  & 8.0 & 190.4 & .054 \\
\rowcolor{conavbg}
\quad + memory
  & 39.6\rlap{\,\dgood{(+4.8)}}
  & 47.8 & .389
  & 15.9\rlap{\,\dgood{(+2.9)}}
  & 136.7 & .117
  & 11.6\rlap{\,\dgood{(+3.6)}}
  & 196.5 & .086 \\
\rowcolor{conavbg}
CoNav-UAV (Qwen3-VL-32B)
  & 31.9 & 48.7 & .314
  & 2.9 & 141.8 & .029
  & 13.0 & 180.2 & .097 \\
\rowcolor{conavbg}
\quad + memory
  & \underline{41.3}\rlap{\,\dgood{(+9.4)}}
  & \underline{47.1} & \underline{.402}
  & 6.5\rlap{\,\dgood{(+3.6)}}
  & 139.5 & .060
  & \underline{15.2}\rlap{\,\dgood{(+2.2)}}
  & 188.3 & \underline{.113} \\
\rowcolor{conavbg}
CoNav-UAV (Qwen3-VL-8B)
  & 34.8 & 49.9 & .334
  & 5.8 & 129.0 & .058
  & 10.1 & 199.6 & .094 \\
\rowcolor{conavbg}
\quad + memory
  & 40.6\rlap{\,\dgood{(+5.8)}}
  & 49.5 & \textbf{.405}
  & 6.5\rlap{\,\dgood{(+0.7)}}
  & 131.1 & .057
  & 10.1\rlap{\,\dgood{(+0.0)}}
  & 197.1 & .082 \\
\bottomrule
\end{tabularx}
\end{table}

\begin{table}[!htbp]
\caption{%
  Full per-tier cross-scene results on \texttt{airsim23}\rev{, with zero-shot (ZS) and fine-tuned (FT) baselines and CoNav-UAV without and with the transferred leader memory}.
  Best in \textbf{bold}; second-best \underline{underlined}.
  \rev{Parentheses give the OSR change due to memory, with increases in {\color{teal}green} and decreases in {\color{red}red}.}%
}
\label{tab:gen23_full}
\centering\fontsize{8.5}{10.5}\selectfont
\setlength{\tabcolsep}{3pt}
\renewcommand{\arraystretch}{1.05}
\begin{tabularx}{\textwidth}{@{}l *{3}{QZZ}@{}}
\toprule
& \multicolumn{3}{c}{Easy \rev{(100\,m)}}
& \multicolumn{3}{c}{Medium \rev{(200\,m)}}
& \multicolumn{3}{c}{Hard \rev{(300\,m)}} \\
\cmidrule(lr){2-4}\cmidrule(lr){5-7}\cmidrule(lr){8-10}
Method
  & OSR$\uparrow$ & NE$\downarrow$ & SPL$\uparrow$
  & OSR$\uparrow$ & NE$\downarrow$ & SPL$\uparrow$
  & OSR$\uparrow$ & NE$\downarrow$ & SPL$\uparrow$ \\
\midrule
CMA~\citep{tan2019lxmert}~{\scriptsize(ZS)}
  & 9.7 & 79.8 & .089
  & 3.0 & 146.3 & .026
  & 0.7 & 236.8 & .007 \\
CMA~\citep{tan2019lxmert}~{\scriptsize(FT)}
  & 12.0 & 74.6 & .110
  & 3.3 & 137.8 & .033
  & 3.0 & 229.3 & .030 \\
\addlinespace[3pt]
Seq2Seq~\citep{sutskever2014seq2seq}~{\scriptsize(ZS)}
  & 6.2 & 81.4 & .062
  & 3.0 & 155.9 & .030
  & 1.7 & 242.0 & .017 \\
Seq2Seq~\citep{sutskever2014seq2seq}~{\scriptsize(FT)}
  & 6.7 & 75.1 & .067
  & 5.3 & 148.0 & .053
  & 0.9 & 219.6 & .008 \\
\addlinespace[3pt]
OpenFly~\citep{openfly2025}~{\scriptsize(ZS)}
  & 6.1 & 87.8 & .061
  & 0.8 & 172.0 & .008
  & 0.0 & 250.4 & .000 \\
OpenFly~\citep{openfly2025}~{\scriptsize(FT)}
  & 11.8 & 75.7 & .115
  & 4.4 & 159.6 & .044
  & 2.8 & 235.3 & .028 \\
\addlinespace[3pt]
AerialVLA~\citep{xu2026aerialvla}~{\scriptsize(ZS)}
  & 19.6 & 69.9 & .192
  & 0 & 171.3 & 0
  & 0 & 256.4 & 0 \\
AerialVLA~\citep{xu2026aerialvla}~{\scriptsize(FT)}
  & 21.4 & 67.8 & .210
  & 9.3 & 151.6 & .093
  & 1.9 & 231.1 & .019 \\
\addlinespace[3pt]
AeroDuo~\citep{wu2025aeroduo}~{\scriptsize(ZS)}
  & 18.6 & 71.0 & .163
  & 0.4 & 177.5 & .001
  & 0.9 & 259.1 & .009 \\
AeroDuo~\citep{wu2025aeroduo}~{\scriptsize(FT)}
  & 22.2 & 66.1 & .207
  & 1.9 & 164.0 & .010
  & 3.7 & 235.6 & .032 \\
\midrule
\rowcolor{conavbg}
CoNav-UAV (GPT-5.6-sol)
  & 18.4 & 64.2 & .170
  & 12.1 & 134.3 & .119
  & 7.3 & 179.2 & .061 \\
\rowcolor{conavbg}
\quad + memory
  & 27.7\rlap{\,\dgood{(+9.3)}}
  & \textbf{60.2} & .233
  & \textbf{18.3}\rlap{\,\dgood{(+6.2)}}
  & \underline{126.0} & \textbf{.167}
  & 9.2\rlap{\,\dgood{(+1.9)}}
  & 176.8 & .074 \\
\rowcolor{conavbg}
CoNav-UAV (GPT-5.4-mini)
  & \underline{27.9} & 61.9 & \underline{.253}
  & 8.5 & 134.5 & .065
  & \textbf{12.7} & \underline{171.6} & \textbf{.103} \\
\rowcolor{conavbg}
\quad + memory
  & \textbf{28.5}\rlap{\,\dgood{(+0.6)}}
  & 65.4 & \textbf{.275}
  & \underline{16.7}\rlap{\,\dgood{(+8.2)}}
  & \textbf{122.3} & \underline{.147}
  & 6.9\rlap{\,\dbad{($-$5.8)}}
  & 181.6 & .056 \\
\rowcolor{conavbg}
CoNav-UAV (Qwen3-VL-32B)
  & 20.6 & 71.8 & .186
  & 8.0 & 143.6 & .059
  & 8.2 & \textbf{166.4} & .051 \\
\rowcolor{conavbg}
\quad + memory
  & 24.0\rlap{\,\dgood{(+3.4)}}
  & \underline{61.1} & .219
  & 8.0\rlap{\,\dgood{(+0.0)}}
  & 143.6 & .054
  & \underline{10.6}\rlap{\,\dgood{(+2.4)}}
  & 192.0 & \underline{.093} \\
\rowcolor{conavbg}
CoNav-UAV (Qwen3-VL-8B)
  & 18.0 & 72.7 & .164
  & 7.4 & 145.2 & .060
  & 1.9 & 202.7 & .016 \\
\rowcolor{conavbg}
\quad + memory
  & 19.9\rlap{\,\dgood{(+1.9)}}
  & 67.9 & .171
  & 8.3\rlap{\,\dgood{(+0.9)}}
  & 137.2 & .077
  & 2.2\rlap{\,\dgood{(+0.3)}}
  & 204.7 & .021 \\
\bottomrule
\end{tabularx}
\end{table}

\section{Data Generation Pipeline}
\label{app:pipeline}

\subsection{High-Altitude Annotation and Verification}
\label{app:pipeline_high}

\begin{figure}[!htbp]
\centering
\begin{minipage}[t]{0.45\textwidth}
  \centering
  \includegraphics[width=\linewidth]{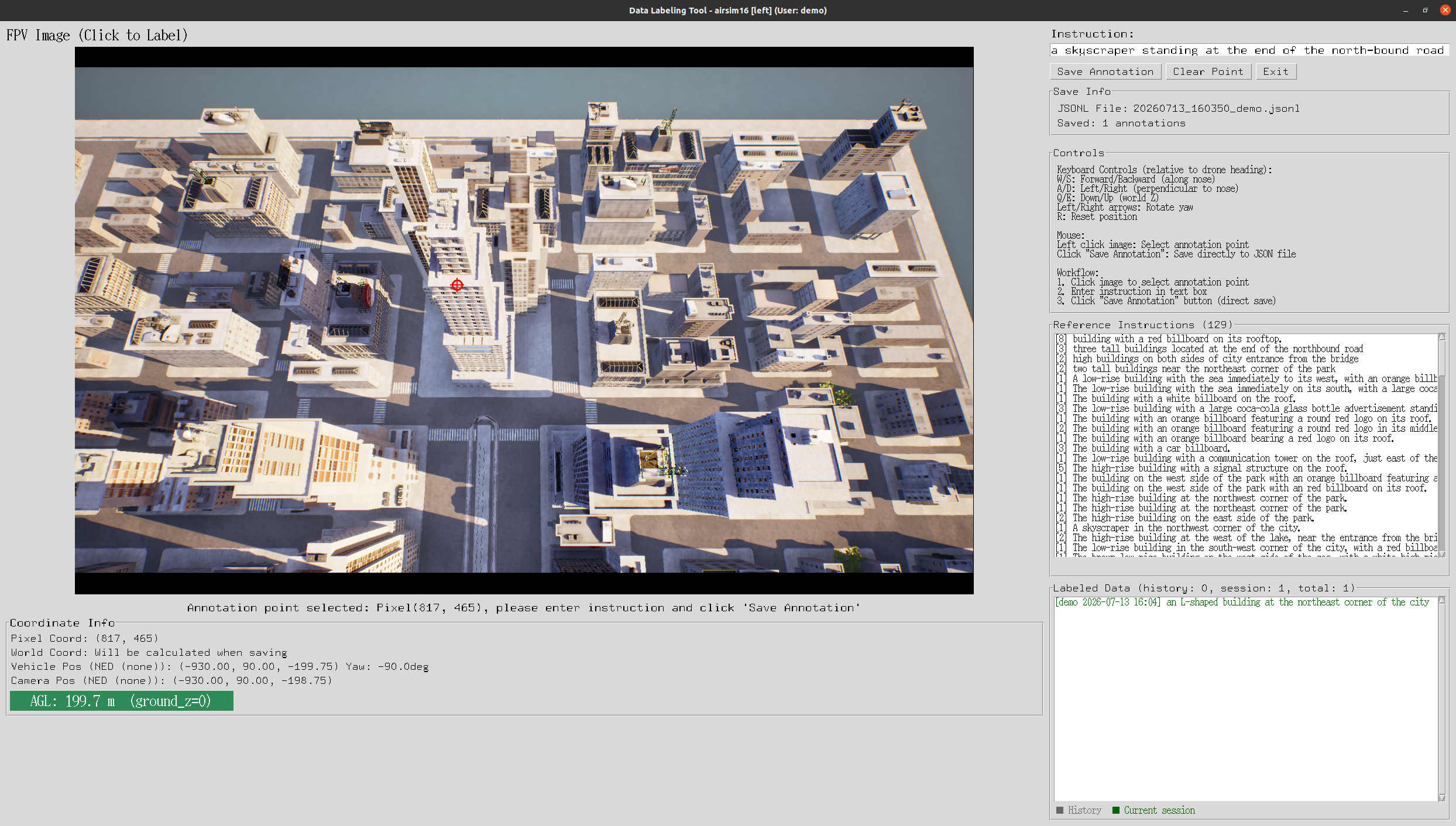}\\
  {\small (a) Annotation interface}
\end{minipage}\hfill
\begin{minipage}[t]{0.45\textwidth}
  \centering
  \includegraphics[width=\linewidth]{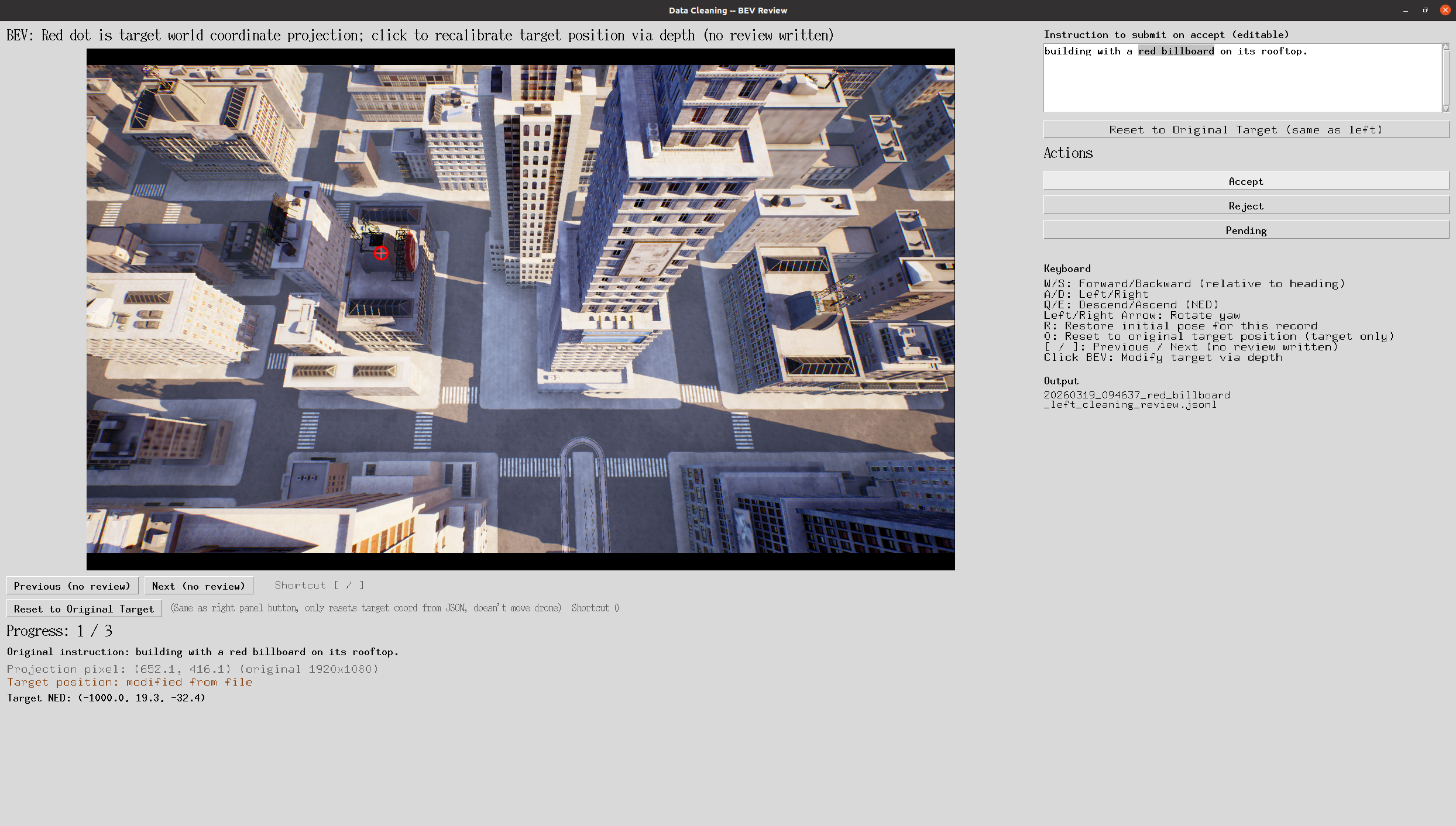}\\
  {\small (b) Verification interface}
\end{minipage}
\caption{%
  Interfaces for high-altitude benchmark construction.
  \textbf{(a)}~The annotator clicks a target in the FPV image;
  the pixel is back-projected through the depth map to world
  coordinates, and a target-oriented instruction is authored.
  \textbf{(b)}~A second annotator verifies each record by
  inspecting the marked target from new viewpoints and accepts,
  rejects, or re-localizes it.%
}
\label{fig:ui}
\end{figure}

Benchmark construction uses two dedicated interfaces
(Fig.~\ref{fig:ui}).

\paragraph{Annotation interface.}
The annotator teleoperates a simulated UAV through the scene in
first-person view and clicks a target object directly in the image.
The clicked pixel is back-projected to world coordinates through
the rendered depth map, combining camera intrinsics with the
current camera pose, so that no manual coordinate entry is
required.
The annotator then composes a target-oriented instruction,
optionally starting from a reference-instruction library.
Each record stores the target world coordinate, the instruction,
the UAV pose at annotation time, a BEV snapshot, and provenance
metadata.

\paragraph{Verification interface.}
A second annotator independently reviews every record.
The interface restores the scene, displays the annotated target as
an overlay marker, and lets the reviewer freely reposition the
camera to inspect the target from new viewpoints.
The reviewer either accepts the record, rejects it, or refines the
target coordinate by re-clicking the object, which triggers the
same depth-based back-projection.
Only accepted records enter the benchmark, and the accept/reject
decision, any coordinate correction, and the review timestamp are
logged for auditability.

\FloatBarrier
\subsection{Low-Altitude Expert Trajectory Synthesis}
\label{app:pipeline_low}

Low-altitude training data is produced fully automatically in four
stages (Fig.~\ref{fig:pipeline_flow}); the resulting corpus is
summarized in Table~\ref{tab:traj_stats}.

\paragraph{Stage 1: geometric twin construction.}
The scene point cloud is voxelized into an occupancy grid, and
building instances are clustered into 2-D contour polygons.
The resulting map preserves collision geometry at a fraction of the
rendering cost and serves as the RL training environment.

\paragraph{Stage 2: start--target pair generation.}
Collision-free spawn points are annotated on the map, and
candidate targets are taken from the building clusters.
Start--target pairs (referred to as \emph{initials}) are then
enumerated and grouped by straight-line distance, with held-out
test pairs excluded from training.
Initials come from three sources depending on the consumer:
\emph{(i)}~bootstrap SFT data pairs random spawn points with random
buildings at 400--600\,m;
\emph{(ii)}~benchmark-target data matches each verified annotation
coordinate to its nearest building cluster and samples spawn
points at the three evaluation tiers, providing expert
trajectories on the benchmark targets for baseline training;
\emph{(iii)}~DAgger data reuses the start--target pairs of the
navigation segments encountered during cooperative rollouts.

\paragraph{Stage 3: PPO expert training and rollout.}
The compact CNN expert described in the main text is trained with
PPO in the geometric twin and then rolled out on the initials,
producing collision-free waypoint sequences.
The expert controls planar displacement only.
Heading is assigned by rule and always points toward the target,
following the intuition that during target approach the obstacles
the UAV must observe and avoid lie predominantly between itself
and the target.
RL is preferred over classical planners for three reasons.
The expert acts on local observations causally consistent with
FPV, preventing the VLA from imitating globally privileged
decisions.
The continuous trade-offs among progress, clearance, and recovery
cannot be captured by fixed heuristics.
Once trained, the expert is an inexpensive, reusable teacher for
arbitrary start--target pairs.

\paragraph{Stage 4: filtering and FPV rendering.}
Rollouts are filtered for spatial diversity by dynamic time
warping (DTW) distance between trajectories and exported as
waypoint files.
Each trajectory is replayed pose-by-pose in the AirSim simulator
at the fixed operating altitude, rendering one FPV image per
waypoint.
Consecutive waypoint differences provide the action labels, so
each trajectory of $N$ waypoints yields $N{-}1$ observation-action
frames.

\begin{figure}[!htbp]
\centering
\begin{tikzpicture}[
  node distance=3mm,
  box/.style={draw, rounded corners=1pt, align=center,
    font=\fontsize{8.5}{10}\selectfont, inner sep=5pt, minimum height=7mm,
    text width=9cm}
]
\node[box] (a) {\textbf{Geometric twin}\\
  point cloud $\to$ occupancy grid + building contours};
\node[box, below=of a] (b) {\textbf{Start--target pairs}\\
  spawn points $\times$ building clusters, distance-banded\\
  (bootstrap / benchmark targets / DAgger segments)};
\node[box, below=of b] (c) {\textbf{PPO expert rollout}\\
  collision-free waypoint sequences};
\node[box, below=of c] (d) {\textbf{Diversity filtering}\\
  waypoint export};
\node[box, below=of d] (e) {\textbf{Simulator replay}\\
  FPV rendering $\to$ observation-action frames};
\draw[-stealth] (a) -- (b);
\draw[-stealth] (b) -- (c);
\draw[-stealth] (c) -- (d);
\draw[-stealth] (d) -- (e);
\end{tikzpicture}
\caption{Low-altitude expert data generation pipeline.}
\label{fig:pipeline_flow}
\end{figure}

\rev{\paragraph{Role of the point cloud.}
CoNav-UAV does not rely on point clouds at deployment, and training the follower requires only instruction-free point-goal trajectories with action labels.
We obtain these labels in a point-cloud twin because it makes the annotation fully automatic.
This is an established practice in aerial navigation, where point-cloud reconstructions support automatic trajectory search~\citep{openfly2025} or serve as the environments in which trajectories are collected~\citep{lee2024citynav}.
Where no reconstruction is available, the same trajectories can be flown by human pilots, and they remain considerably cheaper than conventional VLN data because they involve neither instruction annotation nor target search.}

\section{Leader Grounding Analysis}
\label{app:leader_quality}

We analyze the direct effect of leader updates on target
grounding and compare convergence dynamics across VLMs.
We track dispatched-target precision and GT recall across
iterations (Figure~\ref{fig:dispatch}); both metrics assess only
leader-issued targets and are independent of follower
navigation.

\begin{figure}[!htbp]
\centering
\maxsizebox{\columnwidth}{!}{%
\begin{tikzpicture}
\begin{groupplot}[conav,
  group style={group size=2 by 1, horizontal sep=1.25cm},
  scale only axis, width=5.8cm, height=3.2cm,
  xmin=-0.15, xmax=2.65,
  xtick={0,0.5,1.5,2.5},
  xticklabels={0,.5,1.5,2.5},
  xlabel={Iteration},
  ylabel shift=-4pt,
]
\nextgroupplot[title={Dispatch precision (\%)\,$\uparrow$},
  legend style={at={(1.14,-0.38)}, anchor=north,
    legend columns=3,
    /tikz/every even column/.append style={column sep=4pt}}]
\addplot[cvA, mark=*]
  coordinates {(0,6.7) (0.5,18.3) (1.5,16.0)};
\addplot[cvB, mark=square*]
  coordinates {(0,4.9) (0.5,6.4) (1.5,10.6) (2.5,9.0)};
\addplot[cvC, mark=triangle*, mark size=2.9pt]
  coordinates {(0,4.4) (0.5,8.8) (1.5,9.8)};
\addplot[cvD, mark=diamond*, mark size=2.9pt]
  coordinates {(0,3.7) (0.5,7.4) (1.5,10.8)};
\addplot[cvMuted, densely dashed, mark=o, mark options={solid, draw=cvMuted, fill=white, line width=1.0pt}]
  coordinates {(0,7.2) (0.5,14.7)};
\addplot[cvMuted, densely dotted, mark=x, mark size=2.2pt, mark options={solid, draw=cvMuted, line width=1.0pt}]
  coordinates {(0,8.5) (0.5,9.2)};
\legend{GPT-5.6-sol, GPT-5.4-mini, Qwen3-VL-32B-Instruct, Qwen3-VL-8B-Instruct,
        Claude-Sonnet-4.6, GLM-5V-turbo}
\nextgroupplot[title={GT recall (\%)\,$\uparrow$}]
\addplot[cvA, mark=*]
  coordinates {(0,24.8) (0.5,51.3) (1.5,46.4)};
\addplot[cvB, mark=square*]
  coordinates {(0,16.4) (0.5,22.7) (1.5,46.7) (2.5,42.5)};
\addplot[cvC, mark=triangle*, mark size=2.9pt]
  coordinates {(0,11.2) (0.5,23.9) (1.5,21.5)};
\addplot[cvD, mark=diamond*, mark size=2.9pt]
  coordinates {(0,10.4) (0.5,14.6) (1.5,14.7)};
\addplot[cvMuted, densely dashed, mark=o, mark options={solid, draw=cvMuted, fill=white, line width=1.0pt}]
  coordinates {(0,23.9) (0.5,35.0)};
\addplot[cvMuted, densely dotted, mark=x, mark size=2.2pt, mark options={solid, draw=cvMuted, line width=1.0pt}]
  coordinates {(0,21.6) (0.5,28.7)};
\end{groupplot}
\end{tikzpicture}}
\vspace{-5pt}
\caption{%
  Dispatched-target precision and GT recall (\%) across leader
  iterations.
  Precision is the fraction of dispatched targets near a GT
  target; recall is the fraction of GT targets covered by
  $\ge$1 dispatch.
  Dashed lines denote supplementary leaders (first round
  only).%
}
\label{fig:dispatch}
\end{figure}

First, the gradient-free ICL update converges within one to two
rounds.
Both precision and recall improve substantially in the first one
or two updates.
Dispatch quality then declines mildly, indicating incipient
overfitting, so the evaluated iterations already cover each VLM's
performance peak.

Second, both the convergence rate and the attainable ceiling
depend on backbone capability.
The two Qwen models saturate early at a much lower recall
ceiling than the GPT models, so the benefit of ICL-based
optimization hinges on the backbone's reasoning and long-context
understanding.
The GPT models also converge at different rates.
GPT-5.6-sol explores the scene effectively from the start and
peaks after a single update, whereas GPT-5.4-mini acquires
effective scene exploration in its first update and discovers
higher-quality guidelines only in the second, producing a delayed
recall jump.
Supplementary leaders Claude-Sonnet-4.6 and GLM-5V-turbo
likewise improve after the first round.

Third, the early saturation of Qwen models may stem from
converged memory content or the model's limited ability to
exploit it, and transferring memory banks across VLMs separates
the two (Table~\ref{tab:transfer}, Supp.~\ref{app:leader}).
GPT-5.4-mini reading Qwen iter-1.5 memory exceeds its own
iter-0.5 performance in both precision and recall, so the
Qwen-written memory keeps improving and the bottleneck lies in
the model's ability to exploit it.
VLMs with weaker long-context understanding benefit mainly from
the initial introduction of guidelines and remain insensitive to
further gains in memory quality, which sets the ceiling of
ICL-based optimization.

\rev{A related pattern appears in cross-scene transfer (Table~\ref{tab:memory_ablation}).
Averaged over the two target scenes, memory transfer raises the OSR of GPT-5.6-sol by 4.8\,pp and that of Qwen3-VL-8B by 1.7\,pp, the largest and the smallest gain among the four leaders, in line with the dependence on backbone capability observed above.}

\section{Leader Implementation Details}
\label{app:leader}

\paragraph{Memory transfer experiment.}
Table~\ref{tab:transfer} reports the full memory-transfer
results analyzed in the leader grounding analysis, where
GPT-5.4-mini reads Qwen iter-1.5 memory banks in read-only
mode.

\begin{table}[!htbp]
\caption{%
  Memory transfer: dispatch precision and GT recall when
  GPT-5.4-mini reads Qwen iter\,1.5 memory, compared with GPT's
  own iter\,0.5 and iter\,1.5 memory.%
}
\label{tab:transfer}
\centering\fontsize{8.5}{10.5}\selectfont
\setlength{\tabcolsep}{3pt}
\begin{tabular}{l cc}
\toprule
Memory source & Prec. & Recall \\
\midrule
GPT iter\,0.5 (own) & 6.4\% & 22.7\% \\
GPT iter\,1.5 (own) & 10.6\% & 46.7\% \\
Qwen3-VL-32B-Instruct iter\,1.5 (transfer) & 8.7\% & 36.6\% \\
Qwen3-VL-8B-Instruct iter\,1.5 (transfer) & 8.7\% & 38.9\% \\
\bottomrule
\end{tabular}
\end{table}

\rev{\paragraph{Attribution ablation.}
The intermediate variant in Table~\ref{tab:attribution} isolates the leader's learning from mutual adaptation.
For each test instruction, it reuses the guidelines retrieved in the corresponding iter-2 run of Qwen3-VL-8B with their target-dispatch section removed, and it pairs the leader with the bootstrap follower $\pi_F^{(0)}$.}

\paragraph{Grounding pipeline.}
All evaluated VLMs exhibit limited precise spatial grounding on
aerial BEV imagery.
GroundingDINO therefore first detects candidate objects on each
BEV image, and the VLM filters, confirms, and prioritizes these
candidates through structured reasoning before updating the target
queue.

\paragraph{Action space and budget.}
The leader repositions with fixed-step horizontal translations of
150\,m along the four cardinal directions and discrete altitude
adjustments of 50\,m.
Each episode allows up to 10 grounding rounds, and the leader may
terminate the mission early with \texttt{done}.

\paragraph{Asynchronous scheduling.}
Because VLM inference latency varies across decision steps, the
leader and follower run asynchronously \rev{in our implementation}.
The follower continuously executes toward the current head of the
coordinate queue, and the leader updates the queue whenever its
grounding round completes, so neither agent blocks on the other.

\paragraph{Value-aware retrieval.}
Retrieval uses $k_1{=}5$, $k_2{=}3$, similarity threshold
$\rho{=}0.65$, and $\lambda{=}0.5$.
\rev{New memory entries are initialized with $Q_0^{+}{=}0.5$ or $Q_0^{-}{=}{-}0.2$ (Sec.~\ref{sec:leader}).
The leader reward (Eq.~\ref{eq:rL}) uses the penalty weights $\alpha_{\mathrm{fp}}{=}0.10$, $\alpha_{\mathrm{coll}}{=}0.15$, and $\alpha_{\mathrm{waste}}{=}\alpha_{\mathrm{osc}}{=}0.05$, and is bounded below by $-1$.}
During learning, each retrieved guideline is drawn from the ranked
candidates with probability $0.9$ and uniformly at random from the
memory bank with probability $\varepsilon{=}0.1$, while evaluation
uses $\varepsilon{=}0$.
For cross-scene generalization, the memory bank learned on
\texttt{airsim16} is transferred to the target scenes without any
update.

\section{Follower VLA Fine-Tuning Details}
\label{app:vla}

\paragraph{Model and adaptation.}
The follower is initialized from Qwen2.5-VL-7B-Instruct and
fine-tuned with LoRA ($r{=}32$, $\alpha{=}64$, dropout $0.05$,
applied to all linear layers, with the language-model head and
token embeddings kept trainable).
Training uses bfloat16 with FlashAttention-2 and DeepSpeed ZeRO-2
on 8$\times$\rev{A800}-80GB GPUs.

\paragraph{Action representation.}
Each continuous action $(\Delta x,\Delta y,\Delta\psi)$ is clipped
to the 1\%/99\% percentile range computed on the training corpus
and discretized into 256 bins per dimension, yielding three action
tokens decoded autoregressively.
DAgger batches reuse the original \texttt{airsim16} percentiles so
that the token-to-action mapping stays fixed across iterations.
Expert waypoints are downsampled at interval~3 from the 2\,m PPO
step, bounding per-step follower displacement by 6\,m.

\rev{\paragraph{Sample format and prompt template.}
Each frame forms a single-turn conversation, and training and inference share the template below, in which braces denote fields filled at run time.}

\smallskip
\noindent
\begin{tikzpicture}
\node[draw=black!25, fill=black!3, rounded corners=3pt,
      inner xsep=5pt, inner ysep=5pt, outer sep=0pt,
      text width=\dimexpr\columnwidth-10.8pt\relax, align=left]{%
  \revc\footnotesize
  \setlength{\tabcolsep}{0pt}%
  \renewcommand{\arraystretch}{1.08}%
  \arrayrulecolor{black!18}%
  \begin{tabularx}{\linewidth}{@{}>{\sffamily\bfseries\scriptsize}p{36pt}@{}>{\raggedright\arraybackslash\fontsize{7}{9.2}\selectfont\ttfamily}X@{}}
  System & You are a vision-language-action model. \\[1pt]
  \hline\noalign{\vskip 2pt}
  User & \textrm{\sffamily\scriptsize$\langle$FPV image$\rangle$} \\
       & Instruction: Navigate to the target building. \\
       & Current pose: \{x, y, yaw\} \\
       & Target coordinate: \{x\_t, y\_t\} \\
       & History trajectory: \{[x, y, yaw], ...\} \\[1pt]
  \hline\noalign{\vskip 2pt}
  Assistant & [action\_i][action\_j][action\_k]
  \end{tabularx}%
  \arrayrulecolor{black}};
\end{tikzpicture}\par
\smallskip

\rev{\noindent The user message contains the current FPV image, the current pose, the target coordinate, and a history window of the past 20 poses.
The fixed instruction states the task, and the \texttt{Target coordinate} field specifies the position of the building that the follower must reach.
Poses and coordinates are passed as text, which is native to the VLM and requires no additional state encoder.
The assistant message consists of three action tokens, one per action dimension, and the loss is computed on these tokens only.
Trajectories are split into training and validation sets at the trajectory level with a 90/10 ratio.}

\paragraph{Training schedule.}
All stages share the same optimizer settings, namely a constant
learning rate of $2\times10^{-5}$ with warmup ratio $0.03$,
effective batch size 112, gradient clipping at $1.0$, and
evaluation every 500 steps with the best checkpoint selected by
validation loss.
Stage~1 trains a fresh LoRA on the \texttt{airsim16} corpus
(6{,}516 trajectories, $\sim$615K frames) for 7~epochs.
Each DAgger round continues training the \emph{same} LoRA for
1~additional epoch on the mixture of the initial corpus and all
accumulated DAgger data.
For cross-scene adaptation, a new LoRA is trained on the merged
\texttt{airsim16} weights using target-scene corpora
(1{,}744 trajectories on \texttt{airsim23};
2{,}000 on \texttt{airsim26}).
A from-scratch variant trained on the base model serves solely as
a reference for the transfer analysis below.

\paragraph{Cross-scene transfer.}
Figure~\ref{fig:vla_curves} compares fine-tuning from the
\texttt{airsim16} checkpoint against training from scratch on the
two target scenes.
Warm-starting converges markedly faster, reaching 60--62\% token
accuracy within 500 steps versus 54--55\% for scratch, attains a
lower best validation loss (1.47 vs.\ 1.59--1.62), and maintains a
2--4\,pp token-accuracy advantage throughout training
(64.2\% vs.\ 61.0\% at the final step on \texttt{airsim23};
63.7\% vs.\ 61.0\% on \texttt{airsim26}).
The coordinate-conditioned navigation skill learned on
\texttt{airsim16} therefore transfers across scenes and is not
recovered by scratch training on the moderate target-scene
corpora alone.
This strong transferability is a direct consequence of the role
decomposition in CoNav-UAV.
High-level instruction understanding, cross-modal grounding, and
reasoning are absorbed by the leader, leaving the follower a
responsive, coordinate-conditioned low-level controller.
The resulting skill is largely scene-agnostic, and adapting to a
new scene mainly amounts to accommodating its visual appearance.

\begin{figure}[!htbp]
\centering
\begin{tikzpicture}
\begin{groupplot}[conav,
  group style={group size=2 by 1, horizontal sep=1.5cm},
  width=0.39\textwidth, height=4.0cm,
  xmin=0, xmax=46000,
  xtick={0,10000,20000,30000,40000},
  xticklabels={0,10k,20k,30k,40k},
  scaled x ticks=false,
  xlabel={Training step},
  cycle list={
    {cvA, mark=*},
    {cvA, densely dashed, mark=o, mark options={solid, draw=cvA, fill=white, line width=1.0pt}},
    {cvB, mark=square*},
    {cvB, densely dashed, mark=square, mark options={solid, draw=cvB, fill=white, line width=1.0pt}}},
]
\nextgroupplot[ylabel={Validation loss}, ymin=1.35, ymax=2.9,
  title={(a) Validation loss}, title style={yshift=-0.6ex}]
\addplot coordinates {(500,1.641)(5000,1.465)(10000,1.495)(15000,1.585)
  (20000,1.733)(25000,1.944)(30000,2.236)(35000,2.427)(40000,2.576)(45000,2.730)};
\addplot coordinates {(500,2.252)(5000,1.664)(10000,1.618)(15000,1.621)
  (20000,1.672)(25000,1.769)(30000,1.970)(35000,2.216)(40000,2.424)(45000,2.622)};
\addplot coordinates {(500,1.579)(5000,1.479)(10000,1.512)(15000,1.619)
  (20000,1.748)(25000,1.937)(30000,2.190)(35000,2.424)(40000,2.623)(45000,2.748)};
\addplot coordinates {(500,2.225)(5000,1.669)(10000,1.588)(15000,1.607)
  (20000,1.637)(25000,1.712)(30000,1.836)(35000,2.010)(40000,2.264)(45000,2.496)};
\nextgroupplot[ylabel={Token accuracy}, ymin=0.53, ymax=0.675,
  title={(b) Validation action-token accuracy}, title style={yshift=-0.6ex},
  yticklabel style={/pgf/number format/fixed,
    /pgf/number format/precision=2},
  legend to name=leg:vla, legend columns=2,
  legend style={font=\sffamily\fontsize{8}{9.5}\selectfont, column sep=12pt}]
\addplot coordinates {(500,0.602)(5000,0.646)(10000,0.655)(15000,0.657)
  (20000,0.654)(25000,0.648)(30000,0.640)(35000,0.641)(40000,0.642)(45000,0.642)};
\addplot coordinates {(500,0.544)(5000,0.591)(10000,0.607)(15000,0.617)
  (20000,0.620)(25000,0.620)(30000,0.613)(35000,0.608)(40000,0.609)(45000,0.603)};
\addplot coordinates {(500,0.620)(5000,0.650)(10000,0.657)(15000,0.657)
  (20000,0.651)(25000,0.648)(30000,0.642)(35000,0.637)(40000,0.635)(45000,0.637)};
\addplot coordinates {(500,0.547)(5000,0.592)(10000,0.618)(15000,0.624)
  (20000,0.629)(25000,0.630)(30000,0.628)(35000,0.623)(40000,0.614)(45000,0.610)};
\legend{\texttt{airsim23} fine-tune, \texttt{airsim23} scratch,
        \texttt{airsim26} fine-tune, \texttt{airsim26} scratch}
\end{groupplot}
\end{tikzpicture}\\[3pt]
\pgfplotslegendfromname{leg:vla}
\caption{%
  Cross-scene VLA adaptation: fine-tuning from the
  \texttt{airsim16} checkpoint (solid, filled markers) versus
  training from scratch (dashed, open markers).
  Warm-starting reaches a lower validation loss within the first
  few thousand steps and sustains a consistently higher
  action-token accuracy throughout adaptation.%
}
\label{fig:vla_curves}
\end{figure}

\section{Baseline Implementation Details}
\label{app:baseline_impl}

\paragraph{Seq2Seq.}
Instructions are encoded by a frozen BERT-base followed by a
BiLSTM, and FPV images by a frozen ResNet-50 whose spatial features
are pre-extracted for efficiency.
A pose encoder embeds the current state, a GRU aggregates the
episode context, and a linear head regresses
$(\Delta x,\Delta y,\Delta\psi)$ under a weighted MSE loss
(135.2M parameters, 2.2M trainable).

\paragraph{CMA.}
CMA extends the Seq2Seq backbone with two cross-modal attention
stages, attending from the recurrent state to instruction tokens
and from attended text to spatial visual features, with a second
GRU fusing the attended context
(138.6M parameters, 5.6M trainable).

\paragraph{OpenFly.}
The OpenFly agent follows the OpenVLA recipe, pairing frozen
DINOv2 and SigLIP visual encoders with a LLaMA-2-7B backbone
adapted by LoRA ($r{=}32$, $\alpha{=}64$).
Actions use the same 256-bin tokenization as our follower and are
decoded autoregressively.

\paragraph{AerialVLA.}
Our reproduction instantiates the reactive policy with Qwen3-VL-8B and LoRA.
At each control step, synchronized forward- and downward-facing RGB frames are
resized to $224\times224$ and concatenated vertically, and paired with the
target description and a coarse directional hint derived from the current pose
and target position.
The model autoregressively predicts forward displacement, vertical
displacement, and yaw change, each discretized into 99 bins, together with an
optional \texttt{LAND} token for episode termination.
We adapt the language backbone with LoRA ($r{=}64$, $\alpha{=}128$), train the
visual-language merger modules, and keep the visual encoder frozen.

\paragraph{AeroDuo.}
Our adaptation instantiates the Pilot-LLM with Qwen3-VL-8B and LoRA.
At each high-level step, it receives a $100\,\mathrm{m}\times100\,\mathrm{m}$
RGB BEV crop centered on the follower together with the target instruction.
Learned segmentation queries and a mask head produce a waypoint probability
map, whose high-probability centroid is converted into a global waypoint
without A* post-processing.
The scene-specific low-altitude PPO follower observes a two-channel local
occupancy map derived from the registered point-cloud map and the normalized
relative waypoint, and predicts planar displacement commands of at most
$2\,\mathrm{m}$ for up to 300 steps per waypoint.

\paragraph{Training protocol.}
All baselines share the same two-stage recipe.
They are first pretrained on the converted OpenFly corpus
(Supp.~\ref{app:data_release}), which substantially enlarges
the training data and supplies the generic competence the task
presupposes: grounding an aerial view, avoiding obstacles, and
mapping a described destination to displacement commands.
Since the conversion rewrites route narrations into
target-oriented instructions, this stage already operates on our
input format.
They are then fine-tuned on expert trajectories flown to the
learning-set targets, with start points stratified across the
difficulty tiers, aligning them with the target distribution,
flight altitude, and action statistics of the benchmark.
The two stages are complementary: pretraining provides scale and
transferable skill, while fine-tuning aligns the policy with the
evaluation distribution.

For the cross-scene setting, each baseline starts from its
\texttt{airsim16} checkpoint and is fine-tuned on two sources of
target-scene data: the converted OpenFly trajectories of that
scene, and expert trajectories flown to its annotated
learning-set targets.
The two sources total 4,593 trajectories (145K frames) on
\texttt{airsim23} and 7,498 trajectories (243K frames) on
\texttt{airsim26}.
The CoNav-UAV follower instead consumes only the
coordinate-conditioned subsets of Table~\ref{tab:traj_stats},
1,744 and 2,000 trajectories respectively, which carry no
language annotation, and the leader receives no target-scene
update at all.

\section{Qualitative Examples}
\label{app:qualitative}

\begin{figure}[p]
\centering
\includegraphics[width=0.75\textwidth]{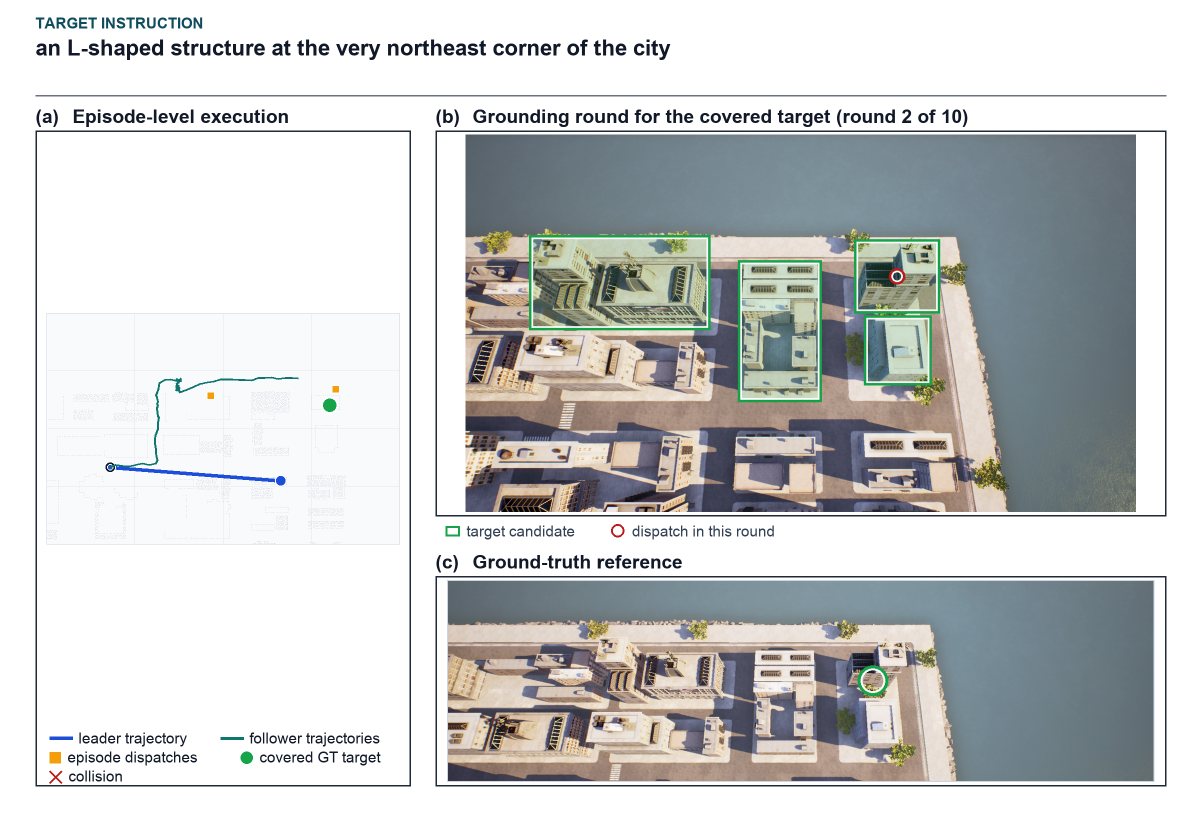}\par
\vspace{0.1em}
\includegraphics[width=0.75\textwidth,trim=0 0 0 38bp,clip]{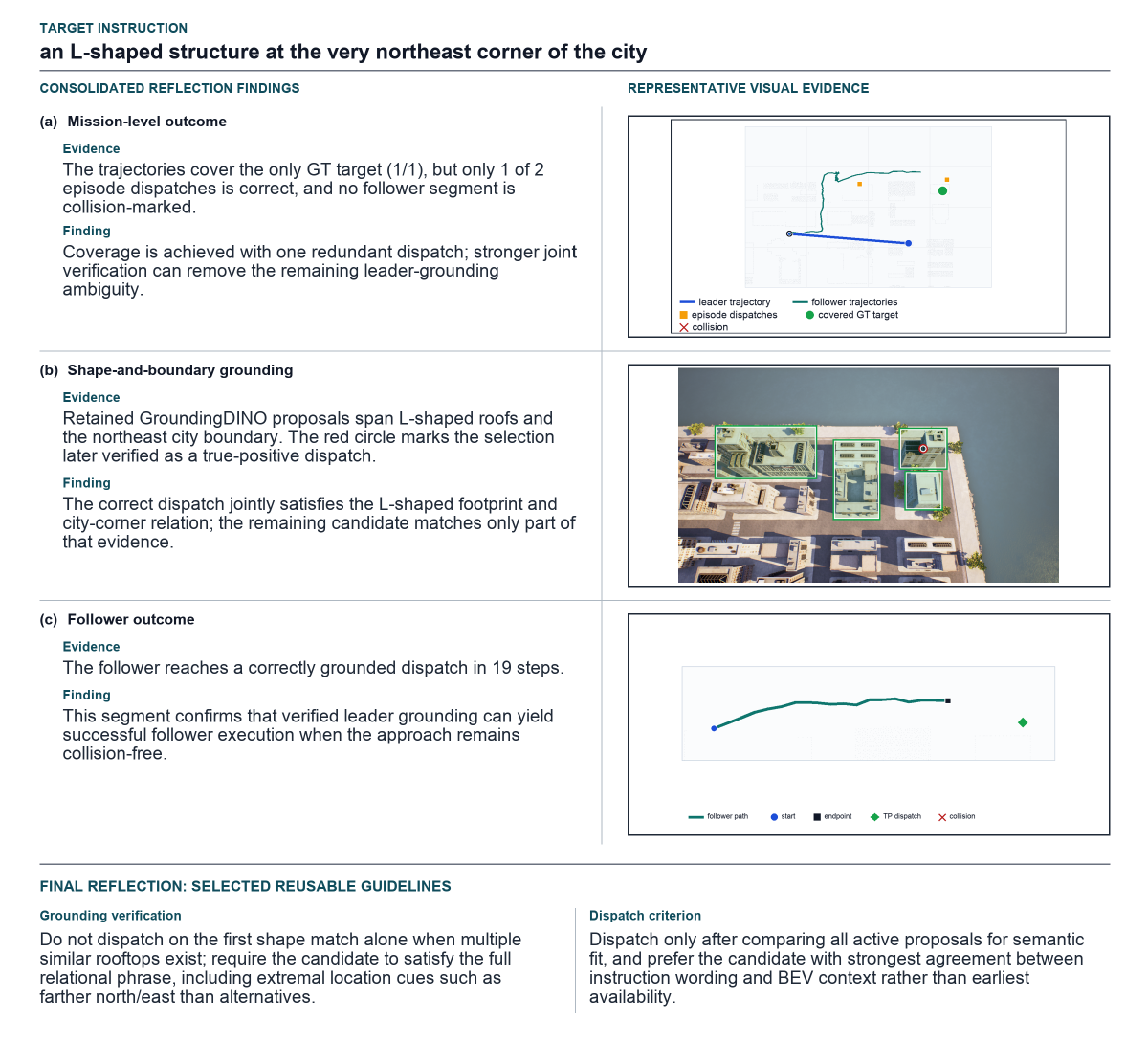}
\caption{Complete-coverage case for an L-shaped structure at the
northeast city corner.  \textbf{Top:} In the full $200\,\mathrm{m}$
episode, the follower covers the sole GT instance without a
collision-marked segment, and one of the two episode dispatches is
semantically correct.  The enlarged covered-target grounding view is
round 2 of 10.  \textbf{Bottom:} The ten-call evidence-grounded
reflection consolidates the multimodal findings and shows two
selected guidelines from its final structured output.}
\label{fig:case_success1}
\end{figure}

\begin{figure}[p]
\centering
\includegraphics[width=0.75\textwidth]{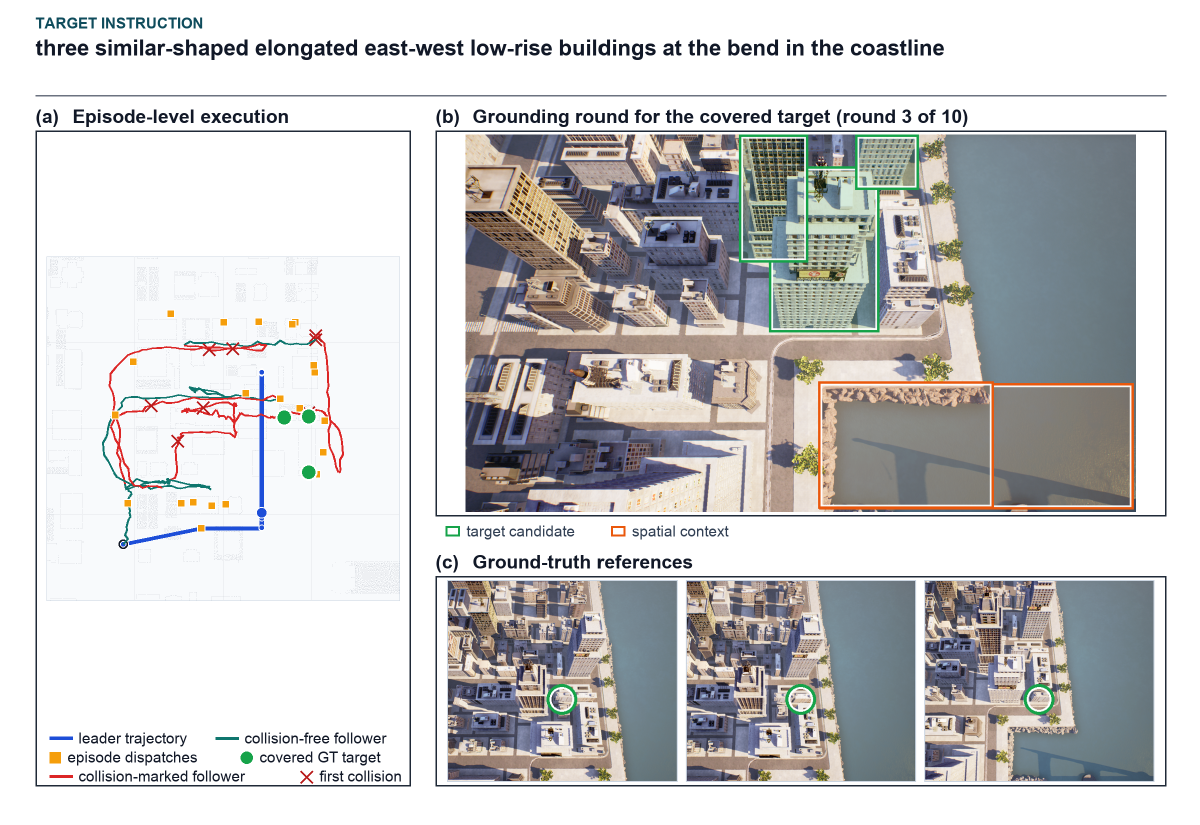}\par
\vspace{0.1em}
\includegraphics[width=0.75\textwidth,trim=0 0 0 38bp,clip]{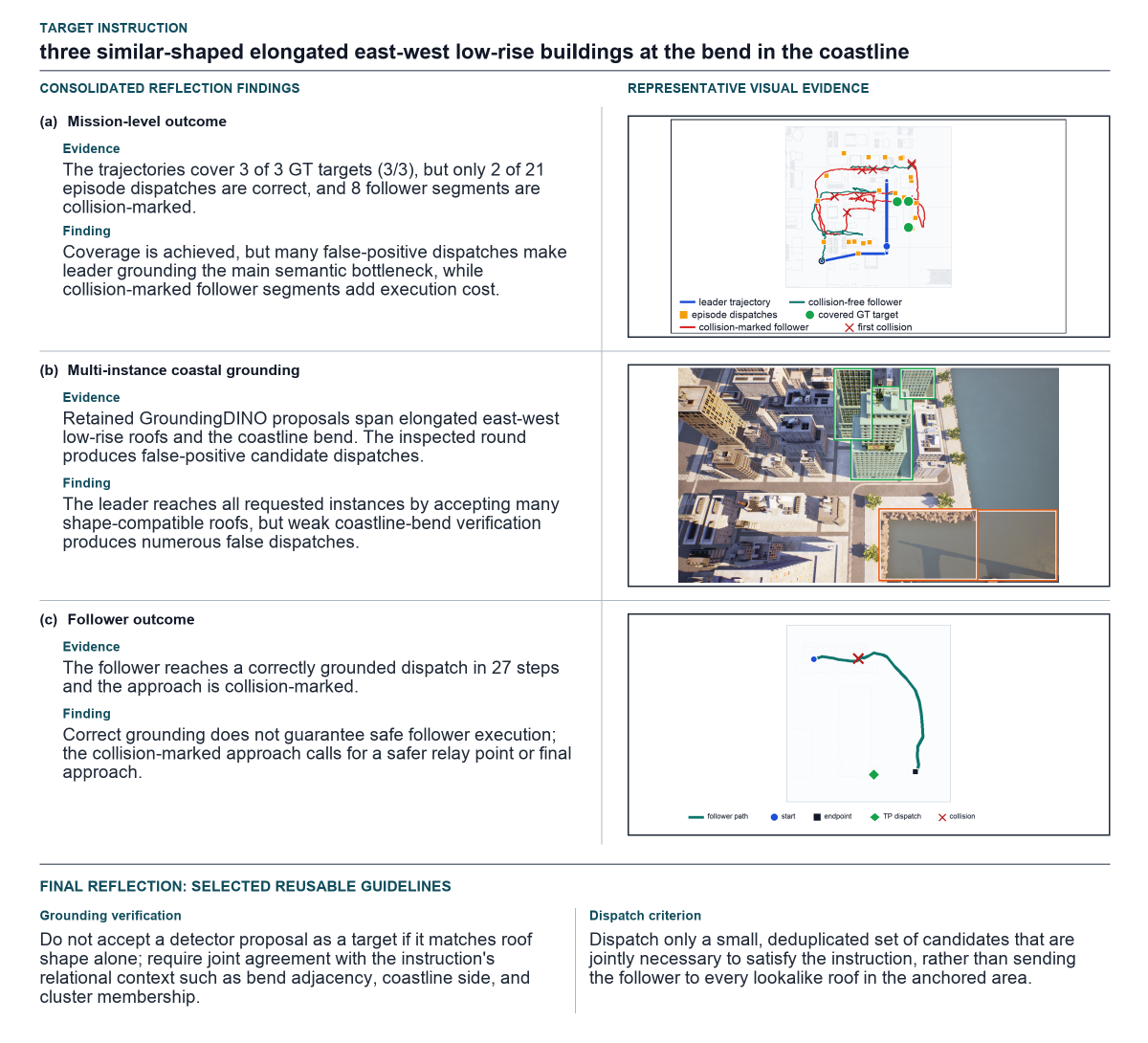}
\caption{High-recall, low-precision case for three elongated
coastal low-rise buildings.  \textbf{Top:} The episode covers all
three GT instances, but only 2 of 21 dispatches are correct and
eight follower segments are collision-marked.  The enlarged
grounding view (round 3 of 10) exposes the broad set of
shape-compatible candidates near the coastline bend.
\textbf{Bottom:} The ten-call reflection diagnoses insufficient
relational verification and over-dispatch, then returns selected
grounding and dispatch guidelines.}
\label{fig:case_broadnet}
\end{figure}

Figures~\ref{fig:case_success1}--\ref{fig:case_broadnet} pair the
execution record and evidence-grounded reflection for two
representative $200\,\mathrm{m}$ search-and-navigation episodes.
The first episode achieves complete GT coverage with two dispatches
and no collision-marked follower segment.  The second exposes a
complementary high-recall, low-precision failure
mode: all GT instances are covered, but broad grounding produces
many false-positive dispatches and repeated collisions.  Together,
the examples make concrete how leader grounding decisions affect
the follower and how the reflector converts episode evidence into
reusable language guidelines.

\clearpage

\clearpage

\subsection{Cooperative Rollouts}
\label{app:qual_rollout}

Each execution summary contains three reader-facing views.  Panel
(a) aggregates the complete episode, including the leader
trajectory, all episode-level dispatches, and the resulting follower
segments.  Panel (b) enlarges one grounding round associated with a
covered target and reports its index among the ten grounding rounds.  Panel (c) provides the
post-mission GT reference used to assess semantic correctness and
coverage.

\subsection{Evidence-Grounded Leader Reflection}
\label{app:qual_reflection}

Reflection is performed after an episode to generate the structured
guideline used by the leader's episodic memory.  For visualization,
we run the same evidence-grounded workflow over the saved episode
artifacts.  In each example, the VLM reflector makes ten function
calls to load context, diagnose textual records, and inspect
grounding, navigation, and BEV evidence.  To keep the figure legible,
context-loading and text-only calls are consolidated into the
left-column findings, while three representative visual inspections
are shown on the right.  \rev{The bottom band reproduces two
reusable guidelines selected from the final structured reflection.}

\section{Data Release and Statistics}
\label{app:data_release}

We will publicly release the following assets.

\paragraph{High-altitude benchmark.}
Verified instruction--target pairs produced by the annotation and
verification interfaces of Supp.~\ref{app:pipeline_high}.
Each item carries an instruction and the coordinates of its GT
targets in the scene's NWU world frame, and annotations sharing
an instruction within a scene are merged into one multi-target
item.
Table~\ref{tab:bench_stats} reports descriptive statistics.
Each scene is split randomly into two thirds of the instructions
for learning and one third for testing, so learning and
evaluation never share an instruction.

\begin{table}[!htbp]
\caption{%
  High-altitude benchmark statistics.
  Multi-target denotes instructions matched by more than one
  ground-truth target; instruction length is counted in words.
  Each scene is split 2/3~learning and 1/3~test, reported as
  instructions/targets.%
}
\label{tab:bench_stats}
\centering\fontsize{8.5}{10.5}\selectfont
\setlength{\tabcolsep}{3pt}
\renewcommand{\arraystretch}{1.1}
\begin{tabular}{lrrrr}
\toprule
& \texttt{airsim16} & \texttt{airsim23} & \texttt{airsim26}
& All \\
\midrule
Instructions            & 93 & 66 & 93 & 252 \\
GT targets              & 132 & 87 & 98 & 317 \\
Multi-target instr.\ & 24 & 9 & 3 & 36 \\
\quad share (\%)        & 25.8 & 13.6 & 3.2 & 14.3 \\
Targets per instr.\ & 1.42 & 1.32 & 1.05 & 1.26 \\
\quad maximum           & 8 & 5 & 4 & 8 \\
Instr.\ length (words) & 14.2 & 13.8 & 15.2 & 14.4 \\
Vocabulary size         & 164 & 207 & 277 & 448 \\
\midrule
Learning (instr./tgt.) & 62/87 & 44/58 & 62/66 & 168/211 \\
Test (instr./tgt.) & 31/45 & 22/29 & 31/32 & 84/106 \\
\bottomrule
\end{tabular}
\end{table}

\begin{figure}[!htbp]
\centering
\includegraphics[width=0.62\textwidth]{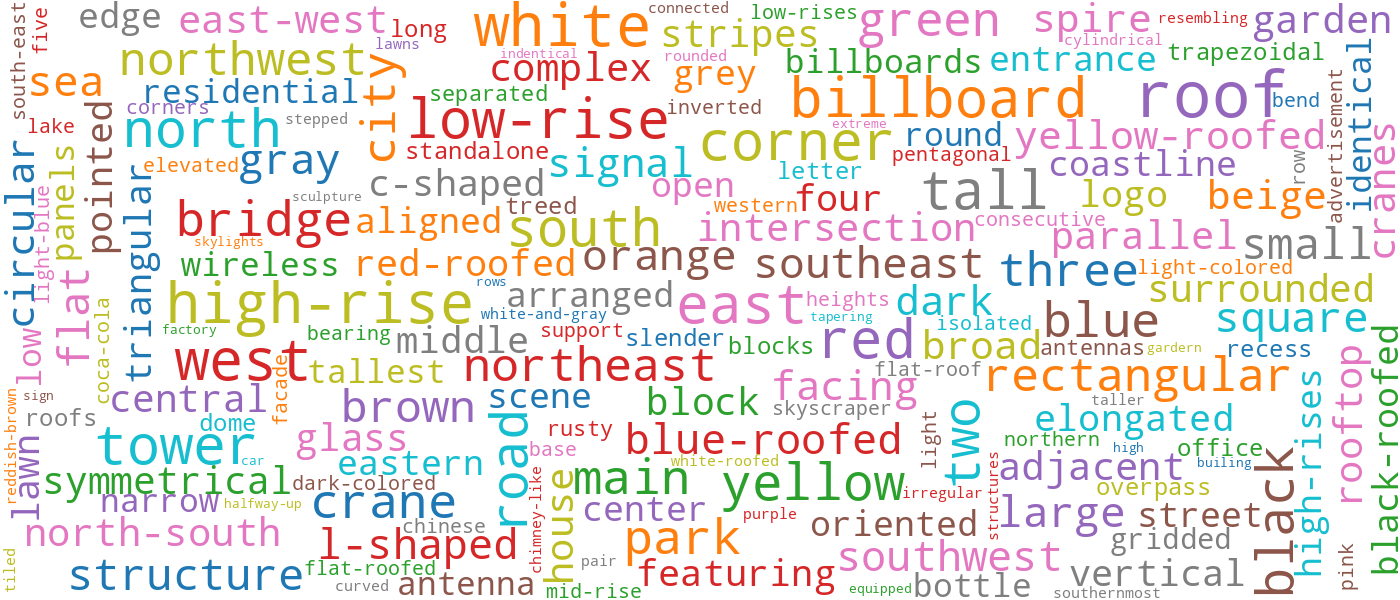}
\caption{%
  Word cloud over all 252 released instructions.
  Descriptions combine appearance attributes (colour, height,
  roof shape, rooftop structures), surrounding landmarks
  (bridge, park, sea, road), and cardinal or relative directions.%
}
\label{fig:wordcloud}
\end{figure}

\begin{figure}[!htbp]
\centering
\begin{tabular}{@{}c@{\;}c@{\;}c@{}}
\includegraphics[width=0.28\textwidth]{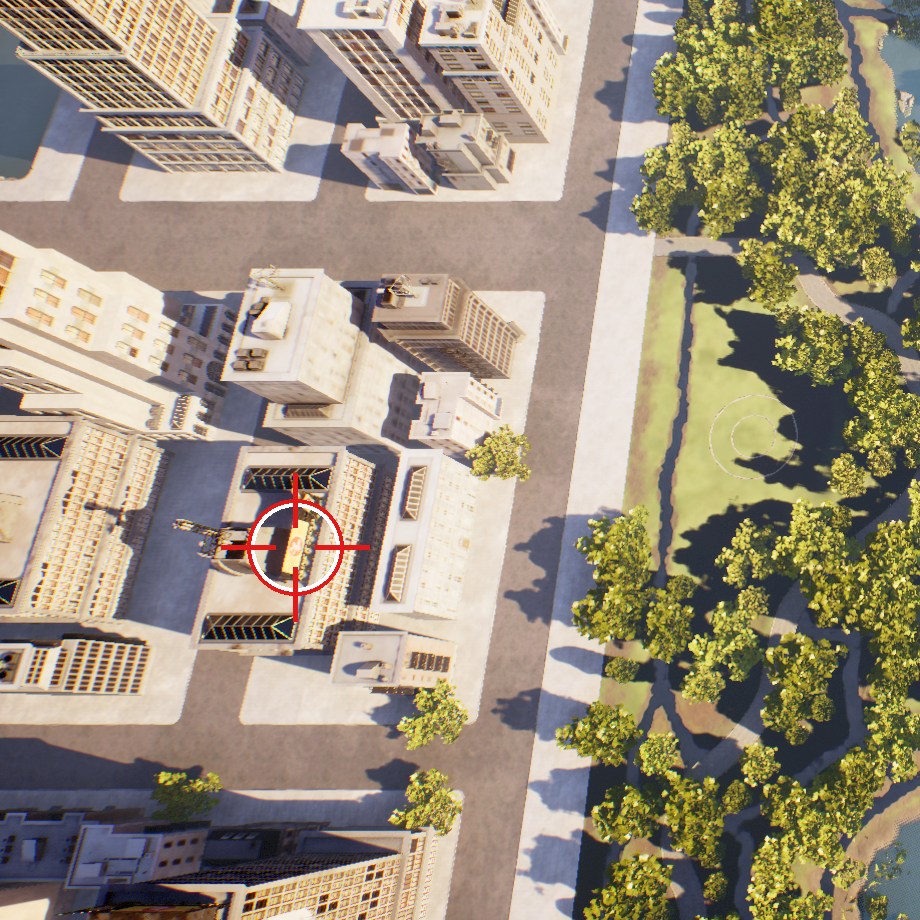} &
\includegraphics[width=0.28\textwidth]{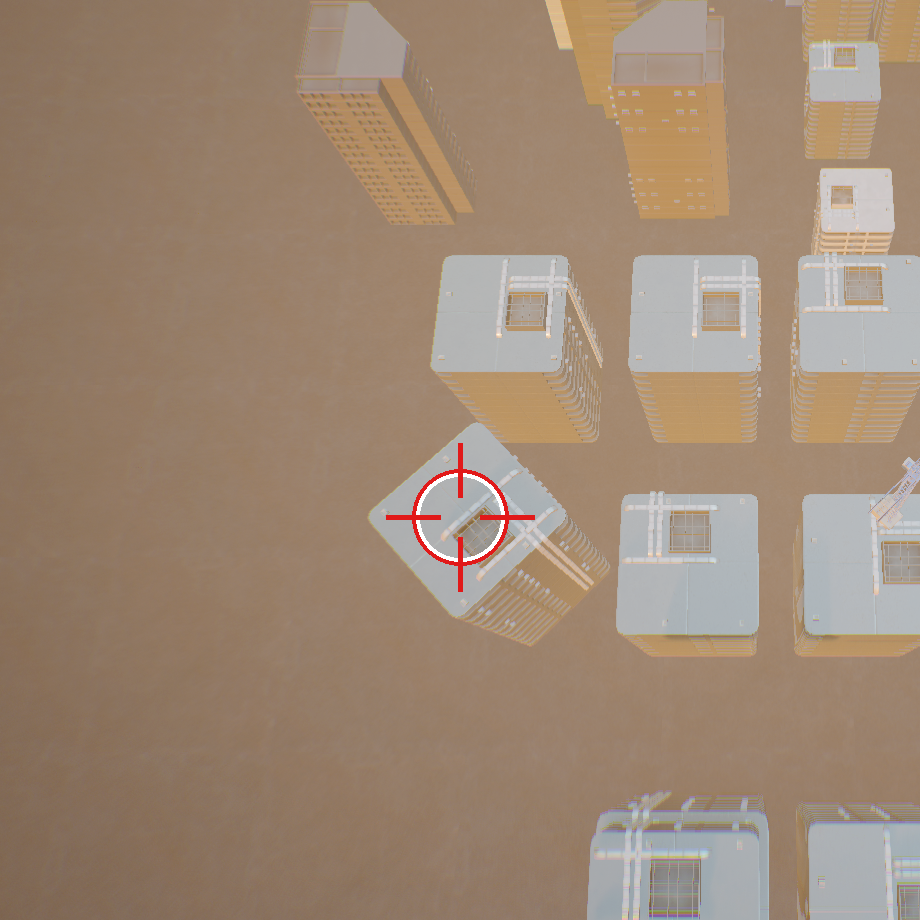} &
\includegraphics[width=0.28\textwidth]{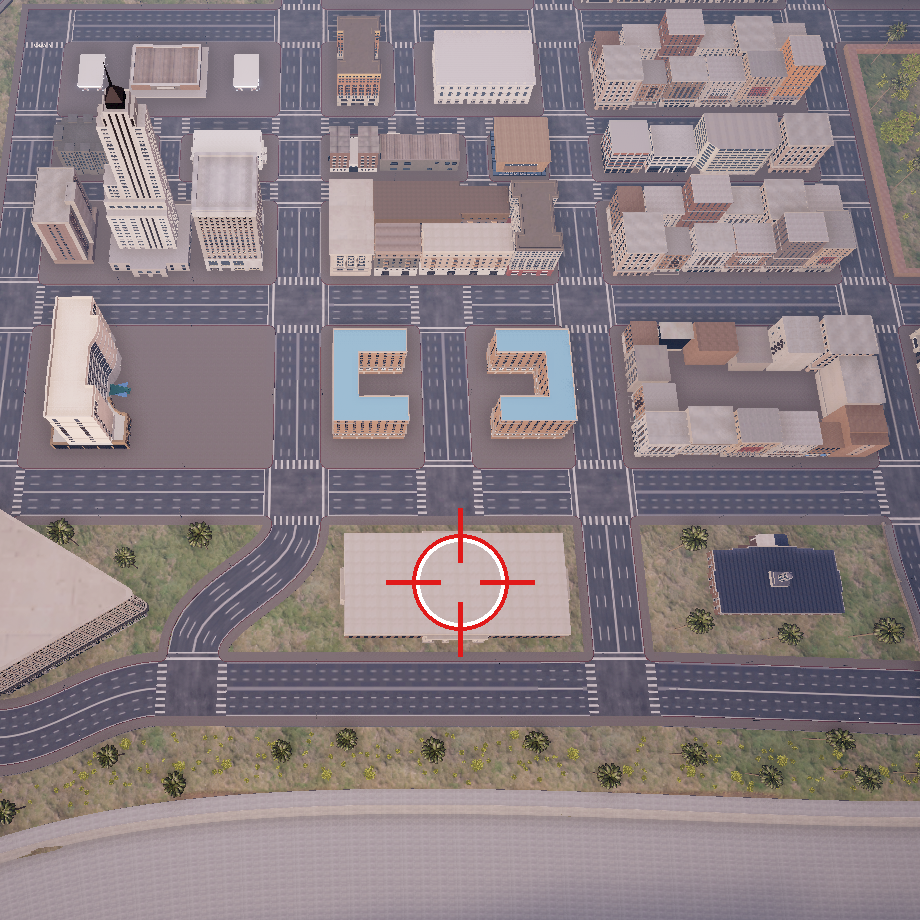} \\
{\scriptsize(a) \texttt{airsim16}} &
{\scriptsize(b) \texttt{airsim23}} &
{\scriptsize(c) \texttt{airsim26}} \\
\end{tabular}
\caption{%
  Annotation examples, with the ground-truth target marked on the
  BEV view used at annotation time.
  \textbf{(a)}~``The building on the west side of the park with an
  orange billboard featuring a round red logo on its roof.''
  \textbf{(b)}~``The diagonally oriented square building with a
  light-blue flat roof, at the southwest corner of the city.''
  \textbf{(c)}~``A white-roofed building with two symmetrical,
  C-shaped blue-roofed buildings to the north and a triangular
  building to the west.''
  Instructions identify a target by appearance, rooftop
  structures, neighbouring buildings, and orientation rather than
  by a route.%
}
\label{fig:ann_examples}
\end{figure}

\paragraph{Low-altitude navigation corpus.}
Coordinate-conditioned expert trajectories produced by the
pipeline of Supp.~\ref{app:pipeline_low}, released as waypoint
sequences with synchronized FPV renderings, all with a waypoint
spacing of about $6\,\mathrm{m}$ (Table~\ref{tab:traj_stats}).
Each scene uses a fixed flight altitude chosen for its building
profile, balancing target observability against
obstacle-avoidance difficulty.
The \texttt{airsim16} subset covers start--target separations of
$400$--$600\,\mathrm{m}$ and bootstraps the follower, and the two
target scenes provide the $\sim$2K-trajectory adaptation subsets
(Sec.~\ref{sec:generalization}).
Owing to its size, this corpus will be released on Hugging Face
upon publication.

\begin{table}[!htbp]
\caption{%
  Low-altitude expert trajectory corpus.
  Frames are consecutive waypoint pairs, each an
  observation-action training sample; altitude is above ground
  level (AGL).%
}
\label{tab:traj_stats}
\centering\fontsize{8.5}{10.5}\selectfont
\setlength{\tabcolsep}{4pt}
\renewcommand{\arraystretch}{1.1}
\begin{tabular}{lrrrr}
\toprule
Scene & Traj. & Frames & Frames/traj. & Altitude (AGL) \\
\midrule
\texttt{airsim16} & 6,516 & 614,857 & 94 & 15\,m \\
\texttt{airsim23} & 1,744 & 142,239 & 82 & 55\,m \\
\texttt{airsim26} & 2,000 & 152,521 & 76 & 20\,m \\
\midrule
Total             & 10,260 & 909,617 & 89 & -- \\
\bottomrule
\end{tabular}
\end{table}

\paragraph{Target-oriented OpenFly conversion.}
OpenFly~\citep{openfly2025} supplies large-scale trajectories
for our scenes, but its instructions narrate turn-by-turn routes,
whereas our task specifies only the destination.
We therefore convert each trajectory into our target-oriented
format for baseline training (Supp.~\ref{app:baseline_impl}),
since the baselines learn instruction understanding jointly with
the fine-tuned policy and require large instruction-conditioned
corpora.
CoNav-UAV itself does not consume this data, as the leader
handles instructions via in-context learning and the follower is
\rev{instruction-free}.
Trajectories with vertical maneuvers are removed.
For each remaining trajectory we render two top-down views from
the point-cloud slice matching its mean flight altitude, a local
crop around the route and a global view of the scene, so the maps
show the geometry actually visible at that level.
A VLM receives both views with the original narration and returns
a concise destination description plus a coarse scene position,
mirroring the appearance-plus-landmark style of our human
annotations.
Grounding the query in rendered geometry keeps spatial relations
verifiable.
Table~\ref{tab:openfly_stats} reports the resulting corpus,
which will likewise be released on Hugging Face upon publication
owing to its size.

\begin{table}[!htbp]
\caption{%
  Target-oriented OpenFly conversion.
  Altitude filtering removes trajectories with vertical maneuvers;
  the remaining trajectories are converted and expanded into
  per-step samples, with a small fraction dropped for degenerate
  paths or failed extraction.%
}
\label{tab:openfly_stats}
\centering\fontsize{8.5}{10.5}\selectfont
\setlength{\tabcolsep}{4pt}
\renewcommand{\arraystretch}{1.1}
\begin{tabular}{lrrrr}
\toprule
& \multicolumn{2}{c}{Original}
& \multicolumn{2}{c}{Converted} \\
\cmidrule(lr){2-3}\cmidrule(lr){4-5}
Scene & Traj. & Alt.-filtered & Traj. & Frames \\
\midrule
\texttt{airsim16} & 9,556 & 8,091 & 6,000 & 211,850 \\
\texttt{airsim23} & 3,429 & 3,429 & 3,000 & 98,201 \\
\texttt{airsim26} & 9,690 & 8,500 & 5,353 & 180,489 \\
\midrule
Total & 22,675 & 20,020 & 14,353 & 490,540 \\
\bottomrule
\end{tabular}
\end{table}

\rev{\section{Toward Physical Deployment}
\label{app:deployment}

\rev{This section reports our preparation for physical deployment}, which covers the choice and compression of the follower backbone, onboard inference, the deployment scheme, and target localization \rev{without privileged} sensing.
Hardware figures refer to the Jetson Orin NX-class edge module selected for both UAVs, which provides 16\,GB of memory and 100\,TOPS.

\paragraph{Follower backbone.}
We compare three follower backbones on a point-goal benchmark of $150$ start--target pairs in \texttt{airsim16}, with 30 pairs per 100\,m band up to 500\,m, a 200-step horizon, a $20$\,m success radius, and a $0.5$\,m collision threshold.
All three are trained on the same \rev{point-goal corpus} with the same inputs and the same 256-bin action tokenization, and each task is executed three times with actions sampled at temperature $0.3$.
The 7B VLA follows the follower architecture of Supp.~\ref{app:vla}, and the 0.8B VLA is distilled from it.
The CNN is a ResNet-18-style network with 12M parameters trained from scratch, which receives the FPV image together with the pose, target coordinate, and pose history as numeric features and predicts the three action tokens with classification heads.
Table~\ref{tab:backbone} reports the success rate (SR), the fraction of tasks solved in all three executions (Pass$_3$), the collision rate (CR), and the timeout rate (TR).

Two observations follow.
First, pretraining matters for long-range navigation.
The CNN is comparable to the VLAs within 200\,m but falls behind beyond this range, where its SR lies between $67.8\%$ and $80.0\%$ against $86.7$--$90.0\%$ for the 7B VLA.
Its collision rate is about twice as high and its Pass$_3$ is $18.7$ points lower, and greedy decoding does not close the gap ($73.6\pm3.2\%$ SR over five runs).
The visual and spatial priors of the pretrained backbone are therefore not recovered by a compact network trained on the same data.
Second, the follower's capability survives compression, as the distilled 0.8B VLA attains $88.0\%$ SR and $82.0\%$ Pass$_3$ with about nine times fewer parameters.}

\begin{table}[!htbp]
\caption{%
  \rev{Follower backbones on the point-goal benchmark in \texttt{airsim16} (three executions per task).
  SR is reported as mean $\pm$ standard deviation over the three executions, and the right block gives SR per start--target distance band.}%
}
\label{tab:backbone}
\centering\fontsize{8.5}{10.5}\selectfont
\setlength{\tabcolsep}{5pt}
{\revc
\begin{tabular}{l c cccc ccccc}
\toprule
& & \multicolumn{4}{c}{Overall (\%)} & \multicolumn{5}{c}{SR by distance band (\%)} \\
\cmidrule(lr){3-6}\cmidrule(lr){7-11}
Follower & Params & SR$\uparrow$ & Pass$_3\uparrow$ & CR$\downarrow$ & TR$\downarrow$
  & 0--100\,m & 100--200\,m & 200--300\,m & 300--400\,m & 400--500\,m \\
\midrule
7B VLA & 7B & $87.6\pm1.4$ & 78.7 & 9.8 & 2.7 & 88.9 & 83.3 & 90.0 & 86.7 & 88.9 \\
0.8B VLA (distilled) & 0.8B & $88.0\pm2.9$ & 82.0 & 8.0 & 4.0 & 92.2 & 81.1 & 88.9 & 90.0 & 87.8 \\
CNN (from scratch) & 12M & $77.6\pm1.7$ & 60.0 & 18.2 & 4.2 & 81.1 & 88.9 & 70.0 & 80.0 & 67.8 \\
\bottomrule
\end{tabular}}
\end{table}

\rev{\paragraph{Onboard inference.}
The 7B VLA exceeds the module's memory, since its BF16 weights alone occupy about $16.6$\,GB, whereas the 0.8B VLA occupies about $2.6$\,GB in FP16.
For latency, Qwen2.5-VL-3B with the follower's input and output lengths (one image, four output tokens) averages $1.107$\,s per step over 30 runs in the module's maximum-performance mode.
Visual encoding and prefill dominate this cost, so latency scales roughly with parameter count.
This gives an estimated $0.28$\,s of inference and about $0.41$\,s of total onboard processing (95th percentile, including image preprocessing and command dispatch) for the 0.8B VLA.

\paragraph{Deployment scheme.}
\rev{The two agents differ in their tolerance of latency, and the scheme places them accordingly.
The follower acts at every control step and runs onboard.
The leader replans only once per grounding round, so in the first stage of physical tests it will run at a ground station, on a locally served VLM or through an API.
The two agents exchange only coordinates, poses, and status messages by design, so the link between them carries less than $10$\,kbps, and a dispatch delayed by $1$\,s costs the follower less than one control step.}

\rev{To avoid hovering while a prediction is computed, the follower will track its waypoints asynchronously, in the manner of carrot-chasing guidance~\citep{sujit2014pathfollowing}.
It will fly toward the most recent waypoint at constant speed during the next inference and apply each prediction relative to the pose at which the corresponding image was captured, so that the motion accumulated during inference does not displace the waypoint.
The waypoints are about $6$\,m apart by design (Supp.~\ref{app:vla}), so one step takes about $1.2$\,s at a cruise speed of $5$\,m/s, roughly three times the estimated onboard processing time.
Each waypoint will thus be available well before the previous one is reached, which permits continuous flight at this speed.}

\paragraph{Target localization without privileged sensing.}
The leader maps a selected BEV pixel to a world coordinate by casting a ray through the pixel, using the camera intrinsics and the UAV pose, and intersecting it with the ground plane.
The procedure needs no depth map, point cloud, or GT target position, so a physical UAV with a calibrated camera and an onboard pose estimate can apply it directly.
\rev{Its error comes from the ground-plane assumption, because a ray aimed at an elevated structure such as a rooftop meets the ground beyond that structure and the resulting coordinate is displaced accordingly.
All our experiments use this procedure, so the reported results already include this error.}
We observe that the follower tolerates \rev{it}.
When a dispatched coordinate falls beside or short of the intended building, the follower typically still approaches the building along the commanded bearing in its FPV view, combining visual evidence with the coordinate.
This tolerance relaxes the localization accuracy required on physical platforms.}

\rev{\section{Scope and Generality of the Dual-UAV Design}
\label{app:scope}

\paragraph{Why the follower is a UAV.}
This work studies aerial search, in which the close-range agent must reach a target wherever it lies in the scene.
A low-altitude UAV meets this requirement because its routes are not bound to the road network and it can enter areas that are inaccessible to ground vehicles, as is typical in disaster rescue and infrastructure inspection.

\paragraph{Other follower embodiments and fleets.}
The formulation does not require the follower to be a UAV.
The two agents interact only through a queue of target coordinates, and the game assumes only that the follower pursues a point-goal utility from its own observations.
A wheeled or legged follower would change the action space, the geometric expert, and the navigation capability to which the leader adapts, whereas the Stackelberg structure and both ISL updates would carry over.
\rev{The formulation also extends to several followers, since the leader's dispatch queue can serve each of them and each best-responds to its own target.
We leave both extensions to future work.}

\paragraph{Relation to air--ground cooperation.}
Existing air--ground systems~\citep{zhang2026agcvln,zeng2026airgroundvln} concentrate on how the two platforms share spatial context, either through a fixed interface or within a single centralized model, and do not treat the platforms as separate agents that learn to adapt to each other (Sec.~\ref{sec:related_vln}).
Our contribution addresses this part and is therefore independent of the follower's embodiment.}

\section{Limitations}
\label{app:limitations}

\rev{\paragraph{Leader grounding is the bottleneck.}
The follower solves a geometric point-goal task with dense action supervision and already succeeds in $87.6\%$ of its point-goal test tasks (Supp.~\ref{app:deployment}).
The leader, in contrast, must ground a briefly described target in a city-scale scene with a frozen VLM that is adapted only in context.
Even the strongest leader reaches at most $18.3\%$ dispatch precision and covers at most $51.3\%$ of the GT targets (Figure~\ref{fig:dispatch}), so cooperative performance is currently limited by the quality of the dispatched coordinates.
\revb{Owing to the inherent difficulty of the task, the leader's grounding also exhibits run-to-run variability, which contributes to the standard deviations in Table~\ref{tab:main}, and improving its robustness is a direction for future work.}
The role division makes this bottleneck separable, since the leader can be improved independently of the follower through fine-tuning for aerial grounding or more effective in-context learning.

\paragraph{Evaluation is conducted in simulation.}
\rev{All experiments are conducted in photorealistic simulation, which is the standard setting for aerial VLN~\citep{liu2023aerialvln,wang2025traveluav,openfly2025,xiao2025uavon,wu2025aeroduo} and ensures that every method is evaluated on identical scenes and episodes.
By design, CoNav-UAV already operates under the sensing and communication conditions of physical UAVs, since both agents rely only on onboard RGB cameras and their own poses and communicate only through a lightweight coordinate queue.
Simulation does not fully reflect the limited computation available on board or the latency that it introduces into the follower's control loop.
We address both in Supp.~\ref{app:deployment}.
The follower is distilled into a 0.8B model that fits the onboard computer at a comparable success rate, and its inference latency is estimated from measurements on the target hardware.
An asynchronous control scheme is then designed to sustain continuous flight under this latency.}}

\end{document}